%% file: llmplan.tex
\documentclass{article}

\usepackage{PRIMEarxiv}

\usepackage[utf8]{inputenc} % allow utf-8 input
\usepackage[T1]{fontenc}    % use 8-bit T1 fonts
\usepackage{hyperref}       % hyperlinks
\usepackage{url}            % simple URL typesetting
\usepackage{booktabs}       % professional-quality tables
\usepackage{listings}
\usepackage{tablefootnote}
\usepackage{multirow}
\usepackage{comment}

\def\mund#1{\smallskip\noindent{\bf #1: }}

\usepackage[table]{xcolor}
\usepackage{booktabs}
\definecolor{backcolour}{rgb}{0.95,0.95,0.92}
\newcommand{\specialcell}[2][c]{%
  \begin{tabular}[#1]{@{}p{12cm}@{}}#2\end{tabular}}
\usepackage{amsfonts}       % blackboard math symbols
\usepackage{nicefrac}       % compact symbols for 1/2, etc.
\usepackage{microtype}      % microtypography
\usepackage{lipsum}
\usepackage{fancyhdr}       % header
\usepackage{graphicx}       % graphics
\graphicspath{{media/}}     % organize your images and other figures under media/ folder
\usepackage{caption}
\title{On the Planning Abilities of Large Language Models\\ (A Critical Investigation with a  Proposed Benchmark)
}

\author{
  Karthik Valmeekam \\
  School of Computing \& AI \\Arizona State University, 
  Tempe.\\
  \texttt{kvalmeek@asu.edu} \\
   \And
     Sarath Sreedharan\thanks{Author was at Arizona State University during part of this work} \\
Department of Computer Science,\\ Colorado State University, Fort Collins.\\
  \texttt{sarath.sreedharan@colostate.edu} \\
   \And
     Matthew Marquez\\
  School of Computing \& AI\\ Arizona State University,
  Tempe.\\
  \texttt{mmarqu22@asu.edu} \\
     \And
     Alberto Olmo\\
  School of Computing \& AI\\ Arizona State University,
  Tempe.\\
  \texttt{aolmoher@asu.edu} \\
     \And
     Subbarao Kambhampati\\
  School of Computing \& AI\\ Arizona State University,
  Tempe.\\
  \texttt{rao@asu.edu} \\
}

\begin{document}
\maketitle

\begin{abstract}
Intrigued by the claims of emergent reasoning capabilities in LLMs trained on general web corpora, in this paper, we set out to investigate their planning capabilities. We aim to evaluate (1) how good LLMs are by themselves in generating and validating simple plans in commonsense planning tasks (of the type that humans are generally quite good at) and 
(2) how good LLMs are in being a source of heuristic guidance for other agents--either AI planners or human planners--in their planning tasks. To investigate these questions in a systematic rather than anecdotal manner, we start by developing a benchmark suite 
based on the kinds of domains employed in the International Planning Competition. On this benchmark, we evaluate LLMs in three modes: \textit{autonomous, heuristic} and \textit{human-in-the-loop}. Our results show that LLM's ability to autonomously generate executable plans is quite meager, averaging only about 3\% success rate. The heuristic and human-in-the-loop modes show slightly more promise. In addition to these results, we also make our benchmark and evaluation tools available to support investigations by research community.
\end{abstract}

% keywords can be removed
% \keywords{First keyword \and Second keyword \and More}

\input{rao-intro.tex}

\section{Related Work}

There have been a few earlier works that looked at the planning capabilities of LLMs. Most of them, such as  \cite{huang2022language,ahn2022can} focus on commonsense domains (e.g. moving things in kitchens) and thus evaluate ``zero shot'' capabilities of LLMs. One issue is that in such cases, it is hard to make a judgment about the correctness of the plan, as there is no accepted world model and the humans often give the benefit of doubt for a plausible--but not actually executable--plan.\footnote{Indeed, a similar temptation to give benefit of doubt on accuracy for plausible completions has led some to think LLMs such as ChatGPT can be used directly for search, leading to rather egregious and amusing results.}
Not surprisingly, the works such as SayCan \cite{ahn2022can}, that actually care about executability, limit themselves to using LLMs in  what we are calling ``heuristic-mode"--with the actions suggested by the LLM being vetted by the underlying sound planner or a reinforcement learner with access to a faithful domain simulator. In our work, in contrast, we assume and allow for the problem and domain model to be specified as part of the prompt--thus allowing us to precisely evaluate the executability and quality of the plans suggested by LLMs. The ability to be conditional on the prompt is also critical for such general systems to be customized for the specific domain of interest. Finally, after our initial study and benchmark were made public, another group did a parallel study that largely corroborates our results on the ineffectiveness of LLMs in finding executable plans \cite{silver2022pddl}.

While this paper focuses on the emergent planning abilities of LLMs not trained specifically on planning tasks, a separate question, that is also receiving attention in the literature, is how well do the underlying sequence completion LLM architectures--specifically transformers--do if trained exclusively on
transition sequences. This is the question handled by works like
decision transformer \cite{chen2021decision} and GATO \cite{reed2022generalist}. Note that there is no \textit{a priori} reason to believe that such direct training on transition sequences doesn't allow the trained transformer to predict plan completions with high accuracy.  Indeed earlier works that used pre-transformer technologies such as word vectors have already shown the viability of such approaches for plan completion \cite{zhuo2020discovering}. One question that is still not settled is whether transformers learn interpretable causal world models or mostly get by with pattern finding abilities.

The idea of developing benchmarks to evaluate emergent properties of LLMs is itself not new. Some prominent existing benchmarks include,  BIG-BENCH \cite{bigbench} and  Coin Flip \cite{wei2022chain}. One could also use datasets like GSM8K \cite{cobbe2021training}, AQUA \cite{ling2017program}, SVAMP \cite{patel2021nlp}, CommonsenseQA \cite{talmor2018commonsenseqa} and StrategyQA \cite{geva2021did} for testing different shallow reasoning abilities. The contribution of our paper is a benchmark and curriculum for evaluating the planning capabilities of LLMs.

\section{Background on Automated Planning}
Given that we are interested in investigating the basic reasoning about actions and change problem, we want to look at the most fundamental planning formalism first, namely the goal-directed deterministic planning problem. Colloquially referred to as {\em classical planning problem}, these problem classes can be mathematically represented by using the tuple $\mathcal{P} = \langle \mathcal{D}, \mathcal{I}, \mathcal{G}\rangle$. $\mathcal{D}$ is referred to as the problem domain, $I$ is the initial state and $G$ is the goal specification. The state-space for the planning problem is defined by the possible truth assignment over the predicates.
The domain is again defined by the tuple $\mathcal{D} = \langle \mathcal{F}, \mathcal{A}\rangle$.  $\mathcal{F}$ corresponds to the set of fluents, i.e., the state variable used to define the state space and each fluent corresponds to a predicate with some arity, and  $\mathcal{A}$ correspond to the set of actions that can be performed as part of the planning problem. Each action $a_i[\mathcal{V}] \in \mathcal{A}$ (where $a_i$ is the operator label and $\mathcal{V}$ is the variable used by the operator and each variable could be mapped to an object), can be further defined by two components, the precondition $prec[\mathcal{V}]$ which describes \textit{when} an action can be executed and the effects $eff[\mathcal{V}]$ which defines \textit{what happens} when an action is executed. We will assume that $prec[\mathcal{V}]$ consists of a set of predicates defined over the variables $\mathcal{V}$. An action is assumed to be executable only if its preconditions are met, i.e, the predicates in the precondition hold in the given state.
% Effect
The effect set $eff[\mathcal{V}]$ is further defined by the tuple $\langle add[\mathcal{V}], del[\mathcal{V}] \rangle$, where $add[\mathcal{V}]$ or add effects is the set of predicates that will be set true by the action and $del[\mathcal{V}]$ or delete effects is the set of predicates that will be set false by the action. 
% Grounding
An action is said to be grounded if we replace each of the variables with an object, else it is referred to as a lifted domain model (we use a similar convention to differentiate between lifted and grounded predicates).
% Example

Below we have provided a snippet of an action description from a popular benchmark problem called Blocksworld for action corresponding to picking up a block.
\begin{lstlisting}
(:action pickup
  :parameters (?ob)
  :precondition (and (clear ?ob) (on-table ?ob) (arm-empty))
  :effect (and (holding ?ob) (not (clear ?ob)) (not (on-table ?ob)) 
               (not (arm-empty))))
\end{lstlisting}
The parameter line provides the possible variables, in this case \textit{?ob}, which can stand for possible blocks. The precondition says that you can only pick up a block if it is clear (i.e. predicate \textit{(clear ?ob)} is true for the block), the block is on the table and the arm is empty. The effects tell you that after you execute the action, the predicate \textit{(holding ?ob)} becomes true and the block will no longer be considered \textit{clear}, and \textit{on-table}. Finally, the arm will no longer be considered \textit{empty}. A solution to a planning problem is called a plan, and corresponds to a sequence of actions that once executed in the initial state would lead to a state where the goal specification is true. The actions may additionally be associated with cost, in these cases, one could also talk about optimal plans, i.e., a plan $\pi$ is called an optimal one if no plan exists that is less costly than $\pi$.

% Allow for typing -- 
The above description presents one of the simpler classes of planning models and can be extended in multiple ways including allowing for object typing (including type hierarchy), more complex forms of preconditions and conditional effects, not to mention supporting richer classes of planning formalisms.
\begin{figure}[ht]
    \centering
    \includegraphics[width=\textwidth]{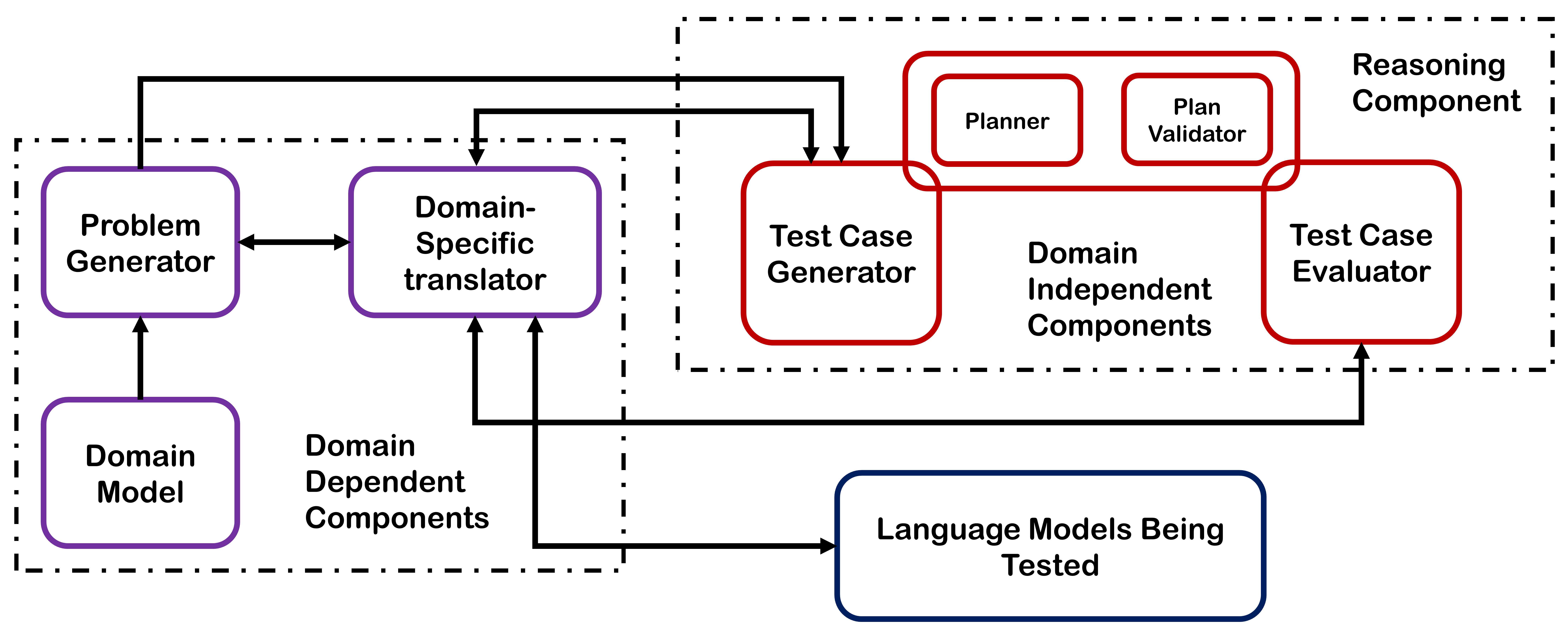}
    % \vspace*{-0.2in}
    \caption{The diagrammatic overview of the overall test framework. Our system consists of a domain-specific component that allows the generation of various instances of the specific PDDL planning problems and the translation from PDDL to text and back. The domain-independent component is responsible for generating the test instances that will be fed into the LLM and verifying the output generated by the LLM.}
    \label{fig:overall}
\end{figure}
\section{Assessment Architecture}

Our basic test framework consists of two categories of components, the domain-independent ones, provided as part of the framework, and the domain-dependent components which need to be developed for each new domain we test.
\subsection{Domain independent component} The domain-independent component is built around a planner and a plan verification component that takes various planning problems and crafts test instances corresponding to various curriculum items. This component provides the mechanism to verify the solutions generated by the LLM. The current method is going to operate almost exclusively on symbolic models (specifically ones specified using PDDL \cite{aeronautiques1998pddl}) and other structured inputs compatible with such representations. The domain-dependent component would be responsible for translating outputs generated by the LLM into forms that can be used by the system.

\subsection{Domain dependent component} The domain-dependent component consists of three main parts. A lifted domain model file, that describes the various actions that may be available to solve any given planning problem, the various predicates that could be used to describe the various relationships over the objects that may be present at a given problem instance of the domain, and the various types of objects that may be part of the given problem. The domain model is lifted because it does not refer to the actual objects that may be part of the problem, but instead, the actions are defined independently of the exact objects it may influence. 

\mund{Problem generator} A planning problem consists of a description of the set of specific objects that are part of the specific planning problem, the initial state (described in terms of the truth values of the various predicates), and a goal description. A valid solution consists of a sequence of actions that can drive the system state to a state that satisfies the goal condition. The role of the problem generator is therefore to generate random problem instances consisting of various objects, initial states, and goals. These problems become the basis of generating the various test cases that we will be using throughout the framework. Any distributional requirements we hope to use in the tests could be built into this problem generator.

\mund{Translator} 
The translator converts the symbolic model information to natural language text and {\em vice versa }. In particular, we are interested in developing a mechanism to translate state information and plans into natural language descriptions similar to what would be provided to humans, thereby normalizing comparison between human planners and LLM planners. For the current testbed (described below), we developed a template-based mechanism to achieve this. In particular, we provide a natural language template for each predicate and each action, and we form texts of states and plans by concatenating these individual strings. In terms of parsing natural language text back into structured forms, the particular task we are interested in is converting plans generated by the LLM back into plan forms that can be used by plan validator tools like \cite{howey2004val}. Since we use our prompts to shape the LLM's output, we require each action in the plan to be listed on a different line. Then, we can parse the exact action and arguments of the action by either using template-based matching or by assuming that the verb in the sentence corresponds to the action and each noun corresponds to an object which forms a parameter of the action (then mapping it to a possible action). 

The domain-independent component is responsible for generating the content for the various prompts that would be generated as part of the different test cases and for validating the output generated by the LLM. As discussed earlier, the component primarily works on formal representations of the problems, so it relies on the translator component to convert any information it generates to natural language or to convert natural language information back to formal representations.
For each test case, we mainly rely on a domain-independent planner and a plan validator to generate the relevant information or to validate the output provided by the LLM. In each case, there is a test-case-specific component that uses the problems provided by the problem generator component to craft specific test-case content. In the next section, we go over each test case and the specific technique we use to generate the contents for the test case.

\section{Current Curriculum for Testing}
%In this section, we go over each specific test case we provide as part of this framework. 
Each test case is meant to evaluate a central reasoning about actions and change capability and is tested in the context of a common sense planning domain. Each test case makes use of the few shot query setting of LLM where the LLM is provided a few sample answers to the specific reasoning ability being tested and is asked to respond to a new instance. The exact form of the prompt will depend on the specific test cases, but every instance will start with a description of the lifted planning domain that describes what actions can be executed, their preconditions and their effects.
The current set of test cases includes the following cases: 
% (1) Plan Generation - Can the LLM come up with valid plans that will achieve a specific goal? (2) Cost Optimal Planning -  Can the LLM come up with plans that are optimal to achieve a specific goal? (3) Reasoning about plan execution - Can the LLM reason about what happens when a plan is executed? (4) Robustness to goal reformulation - Can the LLM recognize the same goal when specified in different ways? (5) Ability to reuse plans - Can the LLM recognize scenarios where it can reuse part or the whole of the original plan to achieve the new goal? (6) Replanning - Can the LLM replan for cases where an unexpected change is reported? and (7) Plan Generalization - Can the LLM take specific plans, extract underlying procedural patterns and apply them to a new instance?
% \begin{lstlisting}[backgroundcolor = \color{backblue}, numbers=none, framexleftmargin = 0.5em,basicstyle = \ttfamily, columns=fullflexible]
\begin{enumerate}
\item Plan Generation - Can the LLM come up with valid plans that will achieve a specific goal?

\item Cost Optimal Planning -  Can the LLM come up with plans that are optimal to achieve a specific goal?

\item Reasoning about plan execution - Can the LLM reason about what happens when a plan is executed?

\item Robustness to goal reformulation - Can the LLM recognize the same goal when specified in different ways?

\item Ability to reuse plans - Can the LLM recognize scenarios where it can reuse part or the whole of the original plan to achieve the new goal?

\item Replanning - Can the LLM replan for cases where an unexpected change is reported?
\item Plan Generalization - Can the LLM take specific plans, extract underlying procedural patterns and apply them to a new instance?
\end{enumerate}
% \end{lstlisting}
Out of the seven test cases, the first two test cases correspond to actual planning problems (i.e. plan generation and cost-optimal planning) and the rest correspond to simpler auxiliary tasks related to reasoning about action and change. Currently, we ground the test cases in a simple common-sense planning domain, Blocksworld. 
Blocksworld problems generally consist of a set of blocks, for making it closer to a common sense domain identified with unique colors, placed either on a table or on top of other blocks and the goal is to arrange some of these blocks in a stack in a particular order. The general expectation here would be that one can pick up a block if it is clear, i.e., there are no other blocks on top of that block and you can only stack a block on top of another block if it is clear. The choice of this particular domain is motivated by both the fact that this is a simple common sense domain and is a very popular domain in planning literature, that has a long history of being used to demonstrate various planning challenges. The domain description is included in the beginning of every prompt. In the rest of the section, we discuss the structure of the prompt for each of the test cases. We provide an example prompt and the corresponding completion generated by GPT-3 for each of the test cases in the Appendix. 

\subsection{Plan Generation}
% \mund{Plan Generation}
Following the lifted domain description, the prompt consists of a few instances of planning problem descriptions (consisting of a description of the initial state, the goal) and the corresponding plan (which ends with a tag, henceforth referred to as the plan-end tag, that denotes the end of the plan) and finally, we end the prompt with a planning problem description. The text generated by the LLM until the plan-end tag is used as a potential candidate for extracting the plan. If the plan-end tag is missing or if the plan cannot be extracted then we ignore that particular instance in our evaluation.

\subsection{Optimal Planning}
% \mund{Optimal Planning}
The prompt is quite similar to the one used in the earlier test case with a few changes. We modify the lifted domain description by including a statement that associates a cost with each action. To make the concept of action cost better fit into common sense domains, we can map the cost to more common concepts like the time taken for executing the action or the amount of money that needs to be spent to execute an action. In the case of each problem description, before the plan is presented we need to explicitly mention that the plan is trying to minimize cost (which depending on the scenario might correspond to saying that the plan takes the least amount of time or the plan correspond to the cheapest plan). The result generated by the LLM is evaluated similarly to the previous query, but in addition to checking if the plan is valid, we also check if the cost of the plan corresponds to the optimal plan cost.

\subsection{Reasoning about plan execution}
% \mund{Reasoning about plan execution}
Here the objective is not to check whether the LLM can come up with plans, but rather if they can predict the outcome of executing an action. The prompt here again starts with the domain description, but instead of providing planning problems and plans, we provide a state, an action sequence and then questions about the state that would result from executing that action sequence in the provided state. Finally the prompt ends with a new state, a new action sequence, and a question about the resulting state. The LLM is expected to come up with an answer, which is checked by applying a plan executor that will try to identify what state will result from the execution of the current action sequence on the state.

\subsection{Robustness to Goal Reformulation}
% \mund{Robustness to Goal Reformulation}
In this test case, we will see if the LLM can recognize goals it has seen before if they are slightly modified. Here the prompt remains the same as the one used for goal-directed reasoning. However, all the example problems have the same initial state, and the last problem provided has not only the same initial state but also the same goal as the example problem. Here the goal may be obfuscated in a few ways, for example, the goal facts may be reordered or one might only include a subset of the original goal specification (meaning the same plan would still work). We can again use the same evaluation technique as the goal-directed reasoning test case to validate the output.

\subsection{Ability to Reuse Plans}
% \mund{Ability to Reuse Plans}
In this test case, we are interested in seeing if the LLM can reuse plans or parts of plans that it has seen before. The prompt is again the same as the goal-directed reasoning, but the prompt ends with a problem that can be solved by a prefix of a previously seen plan. We again keep the initial state the same across the example problems shown. The evaluation remains the same as the goal-directed reasoning test case.

\subsection{Replanning}
% \mund{Replanning}
Replanning corresponds to the problem where there may be an unexpected event that occurs while executing a plan and the system needs to come up with a new plan in response to the event. Here, we focus on the ability of the LLM to replan when unexpected changes are reported. The prompt here starts with a domain description, then a set of instances where an unexpected event occurred during execution, and a new plan in response to the event. In each instance, a planning problem and a corresponding plan are provided at the beginning, the execution of the plan is described and then an unexpected event is noted (event corresponds to some facts unexpectedly turning true or false) and then a new plan from the changed state is presented. The prompt ends with a new case where the plan after replanning is left out and the LLM is expected to complete. The evaluation involves checking whether the new plan is valid from the changed state. The LLM output is evaluated to be true if the new plan it generates achieves the goals from the unexpectedly changed state. 

For the Blocksworld domain, we constrain the unexpected event to be of a specific type. We execute a random prefix of the plan which ensures that some block is held at the end of that prefix. We then change the resulting state by stacking the held block onto another random block which is clear and make the hand empty. This change is reported and the LLM is asked to replan from the changed state.

% \mund{Plan Generalization}
\subsection{Plan Generalization}
In this test case, we want to evaluate whether LLMs can recognize the underlying pattern in the plans provided in the prompt and reuse it for a new planning problem. The prompt is the same as the goal-directed reasoning case, except that all plans were generated by a fixed program. Here the program may contain loops or conditional statements, but can only solve certain types of problems, that is, the initial state and goals meet certain conditions. Such programs can be thought of as a direct generalization of line plans that we have considered in the rest of the paper \cite{srivastava2011qualitative}. Execution of this program for a specific planning problem generates a sequence of actions. In this case, we will provide some example traces generated from the program and ask LLM to come up with a plan for a new problem that could be solved by it. The evaluation again would be to take the generated plan and see if it is valid for the given problem.

\section{Evaluation}

\subsection{Autonomous mode}

\subsubsection{Evaluation of LLMs on the Blocksworld domain}

\begin{figure*}[ht]
    \centering
    \includegraphics[width=\textwidth]{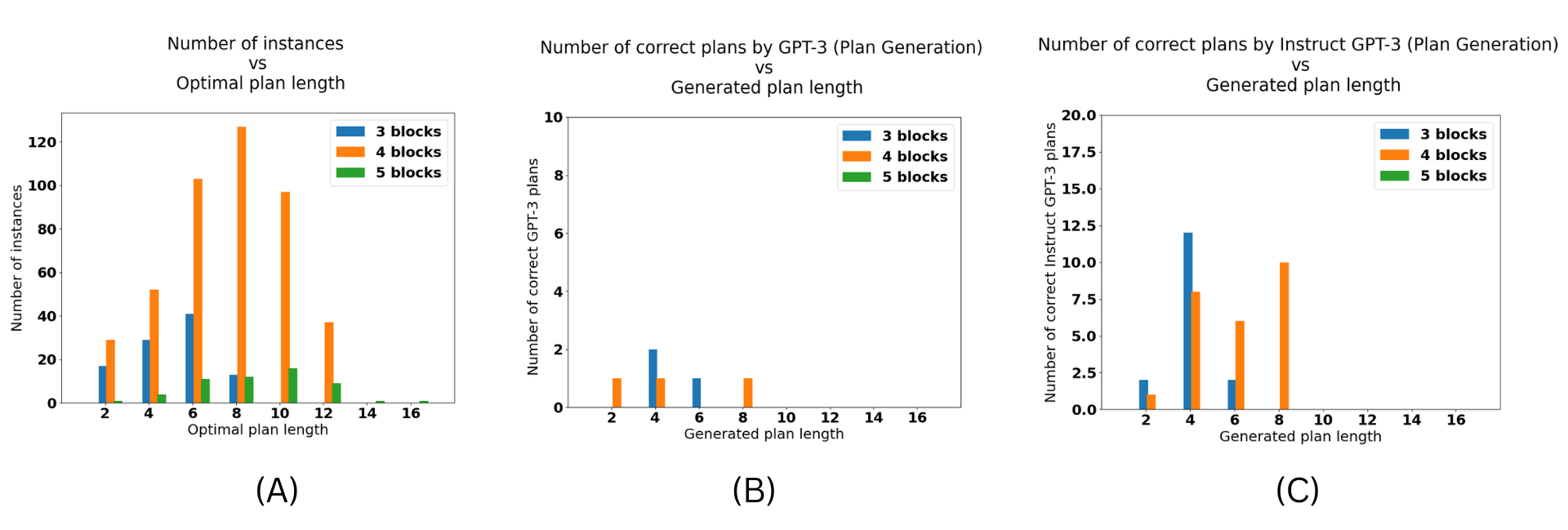}
    % \vspace*{-0.2in}
    \caption{
A detailed comparison of our current dataset, against the instances where GPT-3 or InstructGPT was able to generate a correct plan. The colors correspond to the number of blocks in the blocksworld instance. Note that neither of the LLMs were able to solve an instance which contained five  blocks.
     }
    \label{fig:combined}
\end{figure*}

Our evaluation here primarily focuses on two Large Language Models, GPT-3 and BLOOM. In particular, we evaluated the test framework on the Blocksworld domain.
\begin{table*}[t]
    \centering
    \small
    \begin{tabular}{p{12cm}  p{1.05cm}  p{1.2cm}  p{0.9cm} }
    \toprule
        \textbf{Task} & \multicolumn{3}{c}{
        \centering
        \textbf{Instances correct}} \\ \cmidrule{2-4}
        & GPT-3 & Instruct-GPT3 & BLOOM \\ \midrule[0.08em] 
        \rowcolor{backcolour}
        \specialcell{\textbf{Plan Generation - Preliminary human baseline = 78\%} \\{\scriptsize We showcase an instance and the respective plan as an example and prompt the machine with a new instance. }} & 6/600 (1\%) & 41/600 (6.8\%) & 4/250 (1.6\%) \\ \midrule[0.08em] 
        \rowcolor{backcolour}
        \specialcell{\textbf{Optimal Planning - Preliminary human baseline = 70\%} \\{\scriptsize We showcase an instance, the respective optimal plan and the associated cost as an example and prompt the machine with a new instance. }} & 2/600 (0.3\%) & 35/600 (5.8\%) & 3/150 (2\%)   \\ \midrule[0.08em] 
        \specialcell{\textbf{Replanning} \\{\scriptsize We showcase an instance, the respective plan and present an unexpected change of the state. We then also present a new plan from the changed state. Finally, for a new instance we repeat the same except we ask the machine for the new plan.  }}  & 47/600 (7.8\%) & 40/600 (6.6\%) & 3/100 (3\%) \\ \midrule[0.08em]
        \specialcell{\textbf{Plan Generalization} \\{\scriptsize We showcase an instance and the respective plan as an example and prompt the machine with a new instance. The plans for both the instances can be generated by a fixed program containing loops and conditionals.}} & 33/500 (6.6\%) & 49/500 (9.8\%) & 11/100 (11\%) \\ \midrule[0.08em]
        \specialcell{\textbf{Plan Reuse} \\{\scriptsize We showcase an instance and the respective plan as an example and prompt the machine with an instance which requires only a certain prefix of the plan provided in the example.}} & 0/600 (0\%) & 102/600 (17\%) & 0/100 (0\%) \\ \midrule[0.08em] 
        \specialcell{\textbf{Robustness to Goal Reformulation {\small (Shuffling goal predicates)}} \\{\scriptsize We showcase an instance and the respective plan as an example and prompt the machine with the same instance but shuffle the ordering of the goals.}} & 460/600 (76.6\%) & 467/600 (77.8\%) & 21/100 (21\%) \\ \midrule[0.08em]
        \specialcell{\textbf{Robustness to Goal Reformulation {\small (Full $\rightarrow$ Partial)}}\\{\scriptsize We showcase an instance with a fully specified goal state and the respective plan as an example and prompt the machine with the same instance but provide a partially specified goal state.}}& 407/600 (67.8\%) & 467/600 (77.8\%)   & 9/100 (9\%) \\ \midrule[0.08em] 
        \specialcell{\textbf{Robustness to Goal Reformulation {\small (Partial $\rightarrow$ Full)} }\\ {\scriptsize We showcase an instance with a partially specified goal state and the respective plan as an example and prompt the machine with the same instance but provide a fully specified goal state.}}& 122/600 (20.3\%) & 363/600 (60.5\%) & 5/100 (5\%)   \\ \bottomrule \\
        % $>$Reasoning about Plan Execution & $228/500 = 45.6\%$  \\ \hline 
    \end{tabular}
    \caption{LLM Assessment Suite Results on vanilla GPT-3 (davinci), Instruct-GPT3 (text-davinci-002) and BLOOM (176B model). The tasks in the highlighted rows correspond to actual planning problems while the others correspond to simpler auxiliary planning tasks.}
    \label{tab:my_label}
    % \vspace{-5mm}
\end{table*}
In Table \ref{tab:my_label}, we have presented the results of vanilla GPT-3 (Davinci), Instruct-GPT3 (text-davinci-002), and BLOOM on six of the test cases. For vanilla GPT-3 and Instruct-GPT3, we had tested on 600 instances for all test cases except plan generalization (for which 500 instances were tested) while with BLOOM, we had tested on 250 and 150 instances for plan generation and optimal planning test cases respectively, and 100 instances for the rest of the test cases. The experiments with GPT-3 (both the vanilla and instruct versions) took $\sim$30 minutes for each test case while BLOOM took $\sim$36 hours every 100 instances (on 8 NVIDIA-Quadro RTX 8000 GPUs with 48GBs of memory each).
The best results (within each model) were for the auxiliary goal reformulation test cases. For these three cases, all that was required for the LLM was to repeat the same plan as the one shown in the example. Even then, vanilla GPT-3 and Instruct-GPT-3 failed to do that for some of the instances in the first two cases and the majority of the instances in the third case. BLOOM, on the other hand, was poor in all three cases. Coming to the two test cases that correspond to actual planning problems (plan generation and optimal planning), all three models performed poorly with Instruct-GPT3 performing better than the other two.  Further, we found that for instances where Instruct-GPT3 generated the right plans, when the example plan in the prompt was replaced with another example plan, the accuracy dropped drastically. This suggests that the LLM seems to be  primarily relying on pattern matching (rather than inducing some internal model from the prompts).  We would like to point the reader to the appendix for additional experiments (including fine-tuning and domain disguising). Overall, the performance of these LLMs on our benchmark shows that, as of right now, LLMs are pretty ineffective in autonomously reasoning about actions and change.
The blocksworld instances in our benchmark are somewhat simple as most of them have an optimal plan length of $\leq$8 and a maximum of 5 blocks. Figure \ref{fig:combined}A showcases how the 600 blocksworld instances are distributed over the length of the optimal plan and the number of blocks in the instances.\footnote{We utilized the Fast-Downward planner \cite{helmert2006fast} to come up with optimal plans for these instances and the average time taken by the planner to come up with these plans is 0.149 seconds.} GPT-3 and Instruct GPT3 could not solve any of the 5 block instances and could only generate correct plans which have a length of $<=$8 (as shown in Figure \ref{fig:combined}B \& \ref{fig:combined}C), even though the maximum length among all of the generated plans (correct or incorrect) was 18.
% Detailed insights in appendix

\subsubsection{Human Baseline for the Blocksworld}
%domain}
\label{humanbaseline}
We have previously mentioned that planning tasks on the blocksworld domain are anecdotally simple enough for humans to perform. To establish this and come up with a baseline to compare LLMs performance, we conducted an IRB-approved user study where we asked 50 participants to come up with a plan for a blocksworld instance picked at random, from the set of 500 instances that we used for the evaluation of LLMs. We presented the same domain description as we did for the LLMs and then primed them with an example instance. Further, we provided them with an interface where they had two phases of interaction. In the first phase, they could write up plans by themselves for the given instance and then in the second phase, translate them (by picking the closest action from a list of grounded actions). The translated plans were used in the back-end and were evaluated in an automated fashion\footnote{We had also manually evaluated the plans that they wrote in case they made a mistake during the translation phase}. They went through this procedure first for an example instance (where they were provided with a glimpse of the example solution before using the interface) and then for the actual instance. We provided them with a bonus if they came up with a valid plan. 

Out of the 50 participants, 39 of them (78\%) came up with a valid plan. Along with validity, we also tested the optimality of their plans even though they were not required to come up with an optimal plan. Out of the 39 participants, 35 (89.7\%) participants came up with an optimal plan. These initial results show that the blocksworld domain is a simple enough domain where most humans are able to come up with plans (which are also optimal) while LLMs, on the other hand, showcase subpar performance.

\subsection{Heuristic mode}
As mentioned earlier, instead of using LLMs for generating plans (which they seem to not do so well), we could use LLMs as heuristic guidance to drive sound planners.
In this work, we use a local-search planner LPG \cite{gerevini2002lpg} which generates plans by starting with a seed plan and iteratively repairing flaws until a correct plan is found.  We feed the LLM-generated plan as the initial partial plan for LPG's iterative search. We utilized the plans generated by Instruct-GPT3 on the 600 instances of blocksworld domain as the initial plan that was given to the LPG planner for the corresponding instance. We confirmed that all the plans that were generated by this LLM+LPG combination were valid (which is as expected  given that the underlying planner, LPG, is sound). Given the modest size of the test cases, the solution plans were generated within 7 seconds (including the API call to the LLM). These results show that plans generated by LLMs can be quickly `repaired' by a sound planner to guarantee their correctness.

To get an idea of how far the initial LLM generated plans were from the final correct solutions generated by LPG, we measured the Levenshtein edit distance between them.  
While the default LPG local search doesn't aim to minimize the changes to the suggested plan (there do exist versions of LPG that do this; see \cite{nguyen2012generating})
% \footnote{there do exist versions of LPG that do this; see \cite{nguyen2012generating}}
, the edit distances also give an idea of how partially or approximately correct the original LLM plan is.
%the initial plan generated by the LLM and the final plan generated by LLM+LPG to get an insight into the usefulness of the LLM-generated plans. 
We found that the average Levenshtein distance on the 600 instances was 7.22, while the average length of the final plan was 11.7. This shows that more than 50\% of the final plan might have been due to the edits made by the LPG planner to the initial LLM plan.  
\subsection{Human-in-the-loop mode}
Even though LLMs cannot provide sound plans by themselves, they can still offer their insights as plan suggestions directly to the human-in-the-loop which might potentially guide the user
to the correct plan. After all, this sort of {\em computer supported cooperative work (CSCW)} use case has been the staple of LLM applications. 
%into a useful latent direction. 
We investigated the usefulness of LLMs for human planners by conducting a {\em between-subjects} user study. This user-study is similar in setup to the study described in section ~\ref{humanbaseline} with two key differences. (1) In this user study, there were two independent sets of participants. The first group of participants were not provided with any kind of assistance while coming up with plans (similar to the one in Section~\ref{humanbaseline}) whereas the other group were provided with an LLM generated suggestion that they could make use of. (2) At the end of the study, we have also asked both sets of participants to provide subjective feedback, using NASA-TLX assessment tool, to measure their cognitive load. 
%We utilized the NASA-TLX assessment tool to get feedback and measure the cognitive load of the users. 
Additionally, each participant in the second group had to provide feedback on whether the LLM-generated suggestion was correct when they were presented with the suggestion. We utilized the plans generated by Instruct-GPT3 to provide plan suggestions.

We had 23 participants in the first group, where no assistance was provided, and 22 participants in the second group, where the LLM's plan was provided as a suggestion. Out of the 23 in first group, 17 (i.e., 74\%) of them managed to generate a correct final plan, whereas in the second group, 18 out of 22 (i.e., 82\%) generated a correct plan. While this provides some evidence that LLM generated suggestions were helping the users, the statistical signficance of the findings was not high.
We performed independent-samples t-tests on the time taken to come up with a plan and the cognitive load of the task between the two groups. We set the significance level at $\alpha$=0.05 for both the t-tests and ran them to see if the LLM-assisted group had lesser time taken and cognitive-load. For the t-test on the time taken, we received a statistic value of 0.92 and a p-value of 0.81. For the t-test on the cognitive load, we received a statistic value of 0.78 and a p-value of 0.78. This shows that we can't reject the null hypothesis and thus the difference in the time-taken and cognitive load between the two groups is not statistically significant. Further, 4 out of the 22 participants thought that the LLM suggestion was correct and 2 of them submitted the suggestion itself as the plan. This potentially hints at how these methods could feed into automation bias \cite{cummings2017automation}. Overall, the results showcase that there is a slight improvement in the accuracy of the plans generated when the human planners are assisted by the LLMs but there is no statistically significant difference between the two groups in the time taken and the cognitive load on the human planner.

\section{Conclusion and Future Work}
In this paper, we presented a critical investigation of the planning abilities of large language models (LLMs). To this end, we first provided an extensible benchmark where researchers can evaluate current and future large language models.
%and also a reasoning assessment suite for LLMs that consists of various test cases each evaluating a central aspect of reasoning about actions and change. 
We evaluated the planning abilities of LLMs in three different modes. In the autonomous mode, our results show that even in simple common-sense planning domains where humans could easily come up with plans, current SOTA LLMs like GPT-3 and BLOOM exhibit a dismal performance. 
%However, we do not claim that no LLM system could potentially ever perform effective reasoning about actions and change. Our goal is to establish an extensible benchmark where researchers can evaluate current and future large language models. 
In the heuristic mode, we have seen that plans generated by LLMs can be quickly corrected by sound planners like LPG to guarantee soundness. Finally, in the human in the loop mode, we have seen that human planners are slightly better off with an LLM assisting them as having an LLM as a plan assistant showcased modest improvements in the accuracy of the plans generated by the human-in-the-loop.

We look to improve our assessment suite in multiple ways in the future. We plan to include a modified version of the reasoning about plan execution task to ask questions that require a more descriptive answer and provide automated validations for the answers. 
%We also plan to add an evaluation metric that considers partial correctness of plans.  
This benchmark can be extended to other domains, either to common-sense domains (like Virtual Home \cite{puig2018virtualhome}) or to specialized ones. 
%There are also other potential dirctions that are worth looking at in the context of LLMs and planning.
%\subsection{Chain of thought prompting}
%Chain of thought is a prompt engineering method that provides intermediate steps that lead to the final output \cite{wei2022chain}. In the context of reasoning, chain of thought prompting can include information capturing either explanation of correctness, control or hierarchical decomposition information or some combination of these. It would be interesting to see the extent to which such information can help the LLMs in coming up with plans.
We have also performed additional experiments including evaluating a version of GPT-3 fine-tuned on blocksworld instances, and evaluating LLMs on disguised blocksworld domains. These experiments are described in the Appendix.
In conclusion, we hope that this benchmark
\footnote{Link to the github repo: \url{ https://github.com/karthikv792/gpt-plan-benchmark}} 
encourages other researchers to test the capabilities of their systems across different LLM models \cite{chowdhery2022palm, du2021glam, smith2022using, zhang2022opt, rae2021scaling, thoppilan2022lamda, hoffmann2022training} and even those that are fine-tuned for such tasks.

\section{Acknowledgements}
Kambhampati’s research is supported by the J.P. Morgan
Faculty Research Award, ONR grants N00014-16-1-2892,
N00014-18-1-2442, N00014-18-1-2840, N00014-9-1-2119,
AFOSR grant FA9550-18-1-0067 and DARPA SAIL-ON
grant W911NF19-2-0006. We also want to thank OpenAI
and Miles Brundage for letting us get early research access to the
GPT-3 API.

 \bibliography{llmplan}
\bibliographystyle{plain}

\newpage
\appendix
\onecolumn
\section{Appendix}

\subsection{Additional Experiments}
\label{addexpts}
In the following sections, we will look at additional experiments that were done only on the test cases that correspond to actual planning problems (which are Plan Generation and Optimal Planning). 

\subsubsection{Fine tuning}
\begin{table}[ht]
    \centering
    \small
    \begin{tabular}{l c}
    \toprule
        \textbf{Task} & \textbf{Instances correct} \\ 
        & Finetuned-GPT3 \\ \midrule[0.08em]
        \textbf{Plan Generation} & 82/500 (16.4\%)  \\ \midrule[0.08em]
        \textbf{Optimal Planning} & 110/500 (22\%)  \\
        \bottomrule
    \end{tabular}
    \caption{Results of Plan Generation and Optimal Planning in the Blocksworld Domain on Finetuned-GPT3}
    \label{finetuning}
\end{table}
Along with testing GPT-3, Instruct-GPT3 and BLOOM, we have also looked at the utility of fine-tuning GPT-3 on the blocksworld domain. We prepared a dataset consisting of the initial state, goal state and the respective plan for 1000 blocksworld instances. These instances were different from our test set of 500 instances. We fine-tuned GPT-3 (davinci) on this dataset (using the default hyperparameters provided by Open-AI and 80-20 data split) and evaluated on the two test-cases which correspond to actual planning problems. Even though the results (in Table \ref{finetuning}) showcase an uptick in the number of successful plans, the overall performance is still around 20\%. This is unsurprising as \cite{zhang2022paradox} point out that language models tend to focus on the inherent statistical features in reasoning problems which affects their performance on such tasks.

\subsubsection{Mystery blocksworld domain}
\begin{table}[ht]
    \centering
    \small
    \begin{tabular}{l l l l}
    \toprule
        \textbf{Task} & \multicolumn{3}{c}{\textbf{Instances correct}} \\ \cmidrule{2-4}
        & GPT-3 & Instruct-GPT3 & BLOOM\\ \midrule[0.08em]
        \textbf{Plan Generation} & 0/600 (0\%) & 7/600 (1.1\%) & 0/50 (0\%) \\ \midrule[0.08em]
        \textbf{Optimal Planning} & 0/600 (0\%) & 8/600 (1.3\%) & 0/50 (0\%) \\
        \bottomrule
    \end{tabular}
    \caption{Results of Plan Generation and Optimal Planning in the Mystery Blocksworld Domain with deceptive disguising on GPT-3, Instruct-GPT3 and BLOOM.}
    \label{mystery}
\end{table}
\begin{table}[ht]
    \centering
    \small
    \begin{tabular}{l l l}
    \toprule
        \textbf{Task} & \multicolumn{2}{c}{\textbf{Instances correct}} \\ \cmidrule{2-3}
        & GPT-3 & Instruct-GPT3 \\ \midrule[0.08em]
        \textbf{Plan Generation} & 1/600 (0.1\%) & 5/600 (0.8\%)  \\ \midrule[0.08em]
        \textbf{Optimal Planning} & 3/600 (0.5\%) & 4/600 (0.6\%)  \\
        \bottomrule
    \end{tabular}
    \caption{Results of Plan Generation and Optimal Planning in the Mystery Blocksworld Domain with randomized disguising on GPT-3, Instruct-GPT3 and BLOOM.}
    \label{random}
\end{table}
Another popular domain used in the planning literature is the Mystery domain created by Drew McDermott. In this domain, the goal is to get cargo items from one place to another with vehicles having fuel availability constraints. But the domain is disguised by changing the names of predicates and actions to unrelated entities. Testing LLMs on such a domain would give us insights into whether LLMs are infact performing abstract reasoning or if they are using any common-sense knowledge (such as the meaning of the action/predicate names) in coming up with plans. The domain can be disguised in two different ways, deceptive or randomized. Deceptive disguising would require using words that have meaning by themselves but are unrelated in terms of cause and effect, thereby deceiving the LLM. Randomized disguising would use random alpha-numeric names to disguise the domain.
For better comparison of results, instead of the original mystery domain, we came up with a mystery domain which disguises the blocksworld domain on which the LLMs have already been evaluated. We used both deceptive and randomized disguising and
have evaluated on Plan Generation and Optimal Planning test cases. For deceptive disguising, we have evaluated GPT-3, Instruct GPT-3 and BLOOM whereas for randomized disguising we have evaluated GPT-3 and Instruct GPT-3. The results (in Table \ref{mystery} and Table \ref{random}) showcase a decrease in the number of successful plans generated by the LLMs in the mystery blocksworld as opposed to the original blocksworld. These results indicate that LLMs might not be reasoning at an abstract level and might be relying on the underlying meanings of the actions/predicates and the relations between them while coming up with plans.

\subsection{Blocksworld Domain Prompts}
\label{prompts}
\subsubsection{Domain description included in the prompts}
\begin{lstlisting}[caption=Blocksworld Domain Description]
========================================
I am playing with a set of blocks where I need to arrange the blocks into stacks. Here are the actions I can do

Pick up a block
Unstack a block from on top of another block
Put down a block
Stack a block on top of another block

I have the following restrictions on my actions:
I can only pick up or unstack one block at a time.
I can only pick up or unstack a block if my hand is empty.
I can only pick up a block if the block is on the table and the block is clear. A block is clear if the block has no other blocks on top of it and if the block is not picked up.
I can only unstack a block from on top of another block if the block I am unstacking was really on top of the other block.
I can only unstack a block from on top of another block if the block I am unstacking is clear.
Once I pick up or unstack a block, I am holding the block.
I can only put down a block that I am holding.
I can only stack a block on top of another block if I am holding the block being stacked.
I can only stack a block on top of another block if the block onto which I am stacking the block is clear.
Once I put down or stack a block, my hand becomes empty.
========================================
\end{lstlisting}
\subsubsection{Domain description included in the prompts for optimal planning}
\begin{lstlisting}[caption=Blocksworld Domain Description for Optimal Planning]
========================================
I am playing with a set of blocks where I need to arrange the blocks into stacks. Here are the actions I can do:

Pick up a block. It takes 1 minute to pick up a block.
Unstack a block from on top of another block. It takes 1 minute to unstack a block from on top of another block.
Put down a block. It takes 1 minute to put down a block.
Stack a block on top of another block. It takes 1 minute to stack a block on top of another block.

I have the following restrictions on my actions:
I can only pick up or unstack one block at a time.
I can only pick up or unstack a block if my hand is empty.
I can only pick up a block if the block is on the table and the block is clear. A block is clear if the block has no other blocks on top of it and if the block is not picked up.
I can only unstack a block from on top of another block if the block I am unstacking was really on top of the other block.
I can only unstack a block from on top of another block if the block I am unstacking is clear.
Once I pick up or unstack a block, I am holding the block.
I can only put down a block that I am holding.
I can only stack a block on top of another block if I am holding the block being stacked.
I can only stack a block on top of another block if the block onto which I am stacking the block is clear.
Once I put down or stack a block, my hand becomes empty.
========================================
\end{lstlisting}

\subsubsection{Example prompts and completion by GPT-3}
Below we present the first instance of our 500 instances for each of the tasks and the corresponding completion by GPT-3 as an example in the blocks world domain.
\newline
\newline
\noindent \textbf{Plan generation}
\begin{lstlisting}[caption=Plan Generation]
[STATEMENT]
As initial conditions I have that, the red block is clear, the blue block is clear, the yellow block is clear, the hand is empty, the blue block is on top of the orange block, the red block is on the table, the orange block is on the table and the yellow block is on the table.
My goal is to have that the orange block is on top of the blue block.

My plan is as follows:

[PLAN]
unstack the blue block from on top of the orange block
put down the blue block
pick up the orange block
stack the orange block on top of the blue block
[PLAN END]

[STATEMENT]
As initial conditions I have that, the red block is clear, the yellow block is clear, the hand is empty, the red block is on top of the blue block, the yellow block is on top of the orange block, the blue block is on the table and the orange block is on the table.
My goal is to have that the orange block is on top of the red block.

My plan is as follows:

[PLAN]
--------- GPT3 response ---------
unstack the yellow block from on top of the orange block
put down the yellow block
pick up the orange block
stack the orange block on top of the red block
[PLAN END]
--------- Extracted plan ---------
(unstack yellow orange)
(put-down yellow)
(pick-up orange)
(stack orange red)

-------- Ground truth plan ---------
(unstack yellow orange)
(put-down yellow)
(pick-up orange)
(stack orange red)

================SUCCESS================
\end{lstlisting}
\noindent \textbf{Optimal Planning}
\begin{lstlisting}[caption=Optimal planning]
[STATEMENT]
As initial conditions I have that, the red block is clear, the blue block is clear, the yellow block is clear, the hand is empty, the blue block is on top of the orange block, the red block is on the table, the orange block is on the table and the yellow block is on the table..
My goal is to have that the orange block is on top of the blue block. I want to minimize the time taken to achieve my goal.
My plan is as follows:

[PLAN]
unstack the blue block from on top of the orange block
put down the blue block
pick up the orange block
stack the orange block on top of the blue block
[PLAN END]
The total time to execute the plan is 4 minutes.

[STATEMENT]
As initial conditions I have that, the red block is clear, the yellow block is clear, the hand is empty, the red block is on top of the blue block, the yellow block is on top of the orange block, the blue block is on the table and the orange block is on the table..
My goal is to have that the orange block is on top of the red block. I want to minimize the time taken to achieve my goal.
My plan is as follows:

[PLAN] 
--------- GPT3 response ---------
unstack the yellow block from on top of the orange block
put down the yellow block
pick up the orange block
stack the orange block on top of the red block
[PLAN END]
The total time to execute the plan is 4 minutes.
--------- Extracted plan ---------
(unstack yellow orange)
(put-down yellow)
(pick-up orange)
(stack orange red)

-------- Ground truth plan ---------
(unstack yellow orange)
(put-down yellow)
(pick-up orange)
(stack orange red)

---------Optimal Plan---------
-----Correct cost output by LLM----
================SUCCESS===============
\end{lstlisting}
\noindent \textbf{Replanning}
\begin{lstlisting}[caption=Replanning when an unexpected event occurs and changes the state.]
[STATEMENT]
As initial conditions I have that, the red block is clear, the blue block is clear, the yellow block is clear, the hand is empty, the blue block is on top of the orange block, the red block is on the table, the orange block is on the table and the yellow block is on the table.
My goal is to have that the orange block is on top of the blue block.
My plan is as follows:

[PLAN]
unstack the blue block from on top of the orange block
put down the blue block
pick up the orange block
stack the orange block on top of the blue block
[PLAN END]

During execution, an unexpected event has occurred.
After executing the action "pick up the orange block" at step 3 in the plan, the following facts unexpectedly became true: 
the orange block is clear,
the hand is empty,
the orange block is on top of the red block.

The following facts became unexpectedly false: 
the red block is clear,
the hand is currently holding orange block.

After re-planning from the new state, the plan is as follows:
[PLAN]
unstack the orange block from on top of the red block
stack the orange block on top of the blue block
[PLAN END]

[STATEMENT]
As initial conditions I have that, the red block is clear, the yellow block is clear, the hand is empty, the red block is on top of the blue block, the yellow block is on top of the orange block, the blue block is on the table and the orange block is on the table.
My goal is to have that the orange block is on top of the red block.
My plan is as follows:

[PLAN]
unstack the yellow block from on top of the orange block
put down the yellow block
pick up the orange block
stack the orange block on top of the red block
[PLAN END]

During execution, an unexpected event has occurred.
After executing the action "pick up the orange block" at step 3 in the plan, the following facts unexpectedly became true: 
the orange block is clear,
the hand is empty,
the orange block is on top of the red block.

The following facts became unexpectedly false: 
the red block is clear,
the hand is currently holding orange block.

After re-planning from the new state, the plan is as follows:
[PLAN]
--------- GPT3 response ---------
unstack the orange block from on top of the red block
put down the orange block
stack the orange block on top of the yellow block
[PLAN END]
--------- Extracted plan ---------
(unstack orange red)
(put-down orange)
(stack orange yellow)

-------- Ground truth plan ---------

=============FAILURE===============
\end{lstlisting}

\noindent \textbf{Plan Generalization}
\begin{lstlisting}[caption=Plan generalization]
[STATEMENT]
As initial conditions I have that, the blue block is clear, the yellow block is clear, the magenta block is clear, the black block is clear, the cyan block is clear, the green block is clear, the violet block is clear, the gold block is clear, the hand is empty, the blue block is on the table, the yellow block is on the table, the magenta block is on the table, the black block is on the table, the cyan block is on the table, the green block is on the table, the violet block is on the table and the gold block is on the table.
My goal is to have that the blue block is on top of the cyan block, the yellow block is on top of the blue block, the magenta block is on top of the green block, the black block is on top of the yellow block, the cyan block is on top of the gold block, the green block is on top of the black block and the violet block is on top of the magenta block.

My plan is as follows:

[PLAN]
pick up the cyan block
stack the cyan block on top of the gold block
pick up the blue block
stack the blue block on top of the cyan block
pick up the yellow block
stack the yellow block on top of the blue block
pick up the black block
stack the black block on top of the yellow block
pick up the green block
stack the green block on top of the black block
pick up the magenta block
stack the magenta block on top of the green block
pick up the violet block
stack the violet block on top of the magenta block
[PLAN END]

[STATEMENT]
As initial conditions I have that, the red block is clear, the blue block is clear, the black block is clear, the cyan block is clear, the green block is clear, the violet block is clear, the silver block is clear, the hand is empty, the red block is on the table, the blue block is on the table, the black block is on the table, the cyan block is on the table, the green block is on the table, the violet block is on the table and the silver block is on the table.
My goal is to have that the red block is on top of the blue block, the blue block is on top of the silver block, the black block is on top of the red block, the cyan block is on top of the green block, the violet block is on top of the black block and the silver block is on top of the cyan block.

My plan is as follows:

[PLAN]
--------- GPT3 response ---------
pick up the red block
stack the red block on top of the blue block
pick up the blue block
stack the blue block on top of the silver block
pick up the black block
stack the black block on top of the red block
pick up the cyan block
stack the cyan block on top of the green block
pick up the violet block
stack the violet block on top of the black block
pick up the silver block
stack the silver block on top of the cyan block
[PLAN END]
--------- Extracted plan ---------
(pick-up red)
(stack red blue)
(pick-up blue)
(stack blue silver)
(pick-up black)
(stack black red)
(pick-up cyan)
(stack cyan green)
(pick-up violet)
(stack violet black)
(pick-up silver)
(stack silver cyan)

-------- Ground truth plan ---------
(pick-up cyan)
(stack cyan green)
(pick-up silver)
(stack silver cyan)
(pick-up blue)
(stack blue silver)
(pick-up red)
(stack red blue)
(pick-up black)
(stack black red)
(pick-up violet)
(stack violet black)

===============FAILURE==============
\end{lstlisting}

\noindent \textbf{Plan Reuse}
\begin{lstlisting}[caption=Using a certain prefix of the plan provided in the example]
[STATEMENT]
As initial conditions I have that, the red block is clear, the blue block is clear, the yellow block is clear, the hand is empty, the blue block is on top of the orange block, the red block is on the table, the orange block is on the table and the yellow block is on the table.
My goal is to have that the orange block is on top of the blue block.
My plan is as follows:

[PLAN]
unstack the blue block from on top of the orange block
put down the blue block
pick up the orange block
stack the orange block on top of the blue block
[PLAN END]
 
[STATEMENT]
As initial conditions I have that, the red block is clear, the blue block is clear, the yellow block is clear, the hand is empty, the blue block is on top of the orange block, the red block is on the table, the orange block is on the table and the yellow block is on the table.
My goal is to have that the red block is clear, the orange block is clear, the yellow block is clear, the hand is currently holding blue block, the red block is on the table, the orange block is on the table and the yellow block is on the table.
My plan is as follows:

[PLAN]
--------- GPT3 response ---------
unstack the blue block from on top of the red block
put down the blue block
pick up the red block
stack the red block on top of the orange block
[PLAN END]

-------- Ground truth plan ---------
(unstack blue orange)

==============FAILURE================
\end{lstlisting}

\noindent \textbf{Robustness to Goal Reformulation (Shuffling goal predicates)}
\begin{lstlisting}[caption=Changing the order of goal predicates]
[STATEMENT]
As initial conditions I have that, the red block is clear, the blue block is clear, the yellow block is clear, the hand is empty, the blue block is on top of the orange block, the red block is on the table, the orange block is on the table and the yellow block is on the table.
My goal is to have that the orange block is on top of the blue block.

My plan is as follows:

[PLAN]
unstack the blue block from on top of the orange block
put down the blue block
pick up the orange block
stack the orange block on top of the blue block
[PLAN END]

[STATEMENT]
As initial conditions I have that, the red block is clear, the blue block is clear, the yellow block is clear, the hand is empty, the blue block is on top of the orange block, the red block is on the table, the orange block is on the table and the yellow block is on the table.
My goal is to have that the orange block is on top of the blue block.

My plan is as follows:

[PLAN]
--------- GPT3 response ---------
unstack the blue block from on top of the orange block
put down the blue block
pick up the orange block
stack the orange block on top of the blue block
[PLAN END]
--------- Extracted plan ---------
(unstack blue orange)
(put-down blue)
(pick-up orange)
(stack orange blue)

-------- Ground truth plan ---------
(unstack blue orange)
(put-down blue)
(pick-up orange)
(stack orange blue)

==== (GOAL ORDERING CHANGE) SUCCESS ====
\end{lstlisting}

\noindent \textbf{Robustness to Goal Reformulation (Full$\rightarrow$Partial)}
\begin{lstlisting}[caption=Providing an example with fully specified goal and asking to complete for the same example with a partially specified goal]
[STATEMENT]
As initial conditions I have that, the red block is clear, the blue block is clear, the yellow block is clear, the hand is empty, the blue block is on top of the orange block, the red block is on the table, the orange block is on the table and the yellow block is on the table.
My goal is to have that the orange block is on top of the blue block, the red block is clear, the red block is on the table, the blue block is on the table, the hand is empty, the yellow block is on the table, the orange block is clear and the yellow block is clear.

My plan is as follows:

[PLAN]
unstack the blue block from on top of the orange block
put down the blue block
pick up the orange block
stack the orange block on top of the blue block
[PLAN END]

[STATEMENT]
As initial conditions I have that, the red block is clear, the blue block is clear, the yellow block is clear, the hand is empty, the blue block is on top of the orange block, the red block is on the table, the orange block is on the table and the yellow block is on the table.
My goal is to have that the orange block is on top of the blue block.

My plan is as follows:

[PLAN]
--------- GPT3 response ---------
unstack the blue block from on top of the orange block
put down the blue block
pick up the orange block
stack the orange block on top of the blue block
[PLAN END]
--------- Extracted plan ---------
(unstack blue orange)
(put-down blue)
(pick-up orange)
(stack orange blue)

-------- Ground truth plan ---------
(unstack blue orange)
(put-down blue)
(pick-up orange)
(stack orange blue)

========= (Full->Partial) SUCCESS =====
\end{lstlisting}

\noindent \textbf{Robustness to Goal Reformulation (Partial$\rightarrow$Full)}
\begin{lstlisting}[caption=Providing an example with partially specified goal and asking to complete for the same example with a fully specified goal]
[STATEMENT]
As initial conditions I have that, the red block is clear, the blue block is clear, the yellow block is clear, the hand is empty, the blue block is on top of the orange block, the red block is on the table, the orange block is on the table and the yellow block is on the table.
My goal is to have that the orange block is on top of the blue block.

My plan is as follows:

[PLAN]
unstack the blue block from on top of the orange block
put down the blue block
pick up the orange block
stack the orange block on top of the blue block
[PLAN END]

[STATEMENT]
As initial conditions I have that, the red block is clear, the blue block is clear, the yellow block is clear, the hand is empty, the blue block is on top of the orange block, the red block is on the table, the orange block is on the table and the yellow block is on the table.
My goal is to have that the orange block is on top of the blue block, the red block is clear, the red block is on the table, the blue block is on the table, the hand is empty, the yellow block is on the table, the orange block is clear and the yellow block is clear.

My plan is as follows:

[PLAN]
--------- GPT3 response ---------
unstack the blue block from on top of the orange block
put down the blue block
pick up the orange block
stack the orange block on top of the blue block
unstack the red block from on top of the orange block
put down the red block
pick up the yellow block
stack the yellow block on top of the red block
[PLAN END]
--------- Extracted plan ---------
(unstack blue orange)
(put-down blue)
(pick-up orange)
(stack orange blue)
(unstack red orange)
(put-down red)
(pick-up yellow)
(stack yellow red)

-------- Ground truth plan ---------
(unstack blue orange)
(put-down blue)
(pick-up orange)
(stack orange blue)

===== (Partial->Full) FAILURE =====
\end{lstlisting}

\subsection{Mystery Blocksworld Domain Prompts}
\subsubsection{Domain description using deceptive disguise}
\begin{lstlisting}[caption=Mystery Blocksworld Domain Description]
I am playing with a set of objects. Here are the actions I can do

   Attack object
   Feast object from another object
   Succumb object
   Overcome object from another object

I have the following restrictions on my actions:
    To perform Attack action, the following facts need to be true: Province object, Planet object, Harmony.
    Once Attack action is performed the following facts will be true: Pain object.
    Once Attack action is performed the following facts will be false: Province object, Planet object, Harmony.
    To perform Succumb action, the following facts need to be true: Pain object.
    Once Succumb action is performed the following facts will be true: Province object, Planet object, Harmony.    
    Once Succumb action is performed the following facts will be false: Pain object.
    To perform Overcome action, the following needs to be true: Province other object, Pain object.
    Once Overcome action is performed the following will be true: Harmony, Pain object, Object Craves other object.
    Once Overcome action is performed the following will be false: Province other object, Pain object.
    To perform Feast action, the following needs to be true: Object Craves other object, Province object, Harmony.
    Once Feast action is performed the following will be true: Pain object, Province other object.
    Once Feast action is performed the following will be false:, Object Craves other object, Province object, Harmony.
\end{lstlisting}
\subsubsection{Domain description using deceptive disguise for optimal planning}
\begin{lstlisting}[caption=Mystery Blocksworld Domain Description]
I am playing with a set of objects. Here are the actions I can do

   Attack object. It takes 1 minute to do the Attack action.
   Feast object from another object. It takes 1 minute to do the Feast action.
   Succumb object. It takes 1 minute to do the Succumb action.
   Overcome object from another object. It takes 1 minute to do the Overcome action.

I have the following restrictions on my actions:
    To perform Attack action, the following facts need to be true: Province object, Planet object, Harmony.
    Once Attack action is performed the following facts will be true: Pain object.
    Once Attack action is performed the following facts will be false: Province object, Planet object, Harmony.
    To perform Succumb action, the following facts need to be true: Pain object.
    Once Succumb action is performed the following facts will be true: Province object, Planet object, Harmony.    
    Once Succumb action is performed the following facts will be false: Pain object.
    To perform Overcome action, the following needs to be true: Province other object, Pain object.
    Once Overcome action is performed the following will be true: Harmony, Pain object, Object Craves other object.
    Once Overcome action is performed the following will be false: Province other object, Pain object.
    To perform Feast action, the following needs to be true: Object Craves other object, Province object, Harmony.
    Once Feast action is performed the following will be true: Pain object, Province other object.
    Once Feast action is performed the following will be false:, Object Craves other object, Province object, Harmony.
\end{lstlisting}
\subsubsection{Example prompts and completion by GPT-3}
Below we present the first instance of our 500 instances for the Plan Generation and Optimal Planning tasks and the corresponding completion by GPT-3 as an example in the mystery blocks world domain.
\newline
\newline
\noindent \textbf{Plan generation}
\begin{lstlisting}[caption=Plan Generation]
[STATEMENT]
As initial conditions I have that, province object a, province object b, province object d, harmony, object b craves object c, planet object a, planet object c and planet object d.
My goal is to have that object c craves object b.

My plan is as follows:

[PLAN]
feast object b from object c
succumb object b
attack object c
overcome object c from object b
[PLAN END]

[STATEMENT]
As initial conditions I have that, province object a, province object d, harmony, object a craves object b, object d craves object c, planet object b and planet object c.
My goal is to have that object c craves object a.

My plan is as follows:

[PLAN]
--------- GPT3 response ---------
attack object a
succumb object a
feast object c from object a
overcome object a from object c
[PLAN END]
--------- Extracted plan ---------
(pick-up object a)
(put-down object a)
(unstack object c object a)
(stack object a object c)

-------- Ground truth plan ---------
(unstack object d object c)
(put-down object d)
(pick-up object c)
(stack object c object a)

================FAILURE================
\end{lstlisting}
\noindent \textbf{Optimal Planning}
\begin{lstlisting}[caption=Optimal planning]
[STATEMENT]
As initial conditions I have that, province object a, province object b, province object d, harmony, object b craves object c, planet object a, planet object c and planet object d..
My goal is to have that object c craves object b. I want to minimize the time taken to achieve my goal.
My plan is as follows:

[PLAN]
feast object b from object c
succumb object b
attack object c
overcome object c from object b
[PLAN END]
The total time to execute the plan is 4 minutes.

[STATEMENT]
As initial conditions I have that, province object a, province object d, harmony, object a craves object b, object d craves object c, planet object b and planet object c..
My goal is to have that object c craves object a. I want to minimize the time taken to achieve my goal.
My plan is as follows:

[PLAN] 
--------- GPT3 response ---------
attack object a
overcome object a from object b
feast object b from object c
succumb object b
[PLAN END]
The total time to execute the plan is 4 minutes.
--------- Extracted plan ---------
(pick-up object a)
(stack object a object b)
(unstack object b object c)
(put-down object b)

-------- Ground truth plan ---------
(unstack object d object c)
(put-down object d)
(pick-up object c)
(stack object c object a)

================FAILURE================
\end{lstlisting}

\subsection{User study details}
We ran the user studies on an online platform Prolific and paid the participants a wage of \$8.12/hour for the human baseline study (described in Section 6.1.2) and \$10.29/hour for the LLM+human user study (described in Section 6.3). 
\subsubsection{Instructions provided to the participants}
\mund{Consent for Study}
The expected time of participation is between 25-35 minutes. You have the right not to answer any question, and to stop participation at any time. On successful completion, you will be eligible to receive \$5-8 for your participation in this study. We will need to record all the responses provided by the participants during the study. Your consent to participate in this study is completely voluntary.
To protect your privacy, responses from participants will never be used individually while compiling or presenting results of the study. The results of this study may be used in reports, presentations, or publications only in an aggregate form. Please enter your prolific id and click continue with the study if you agree to take part in this study.

\mund{Study details for participants receiving LLM assistance}
In this study, you will be coming up with a plan that achieves certain goal conditions given some initial conditions.
\begin{itemize}
    \item A plan is a sequence of actions that achieve certain goals.
\item A domain consists of the actions that can be done and the restrictions on the actions.
\item A problem in the specified domain will consist of the initial conditions and the goal conditions for which a plan is a solution.
\end{itemize}
You will be dealing with the blocksworld domain which consists of playing with a set of blocks where you need to arrange the blocks into stacks.
You will have to come up with a plan for one blocksworld problem. You will have an AI agent that will help you in coming up with plans. This AI agent is not perfect and can make mistakes.
 You get a base bonus of 50 cents.
 \begin{itemize}
\item If you come up with a successful plan your bonus compensation increases by \$1.
\item If your plan is unsuccessful, your bonus compensation decreases by 50 cents.
\item Random plan submissions will be rejected and the bonus compensation would not be provided for such submissions.
\end{itemize}
We recommend you to have a pen and paper to aid you in visualizing the domain whenever required. We will first look at how the blocksworld domain works and what actions can you do.

\mund{Study details for participants not receiving LLM assistance}
In this study, you will be coming up with a plan that achieves certain goal conditions given some initial conditions.
\begin{itemize}
    \item A plan is a sequence of actions that achieve certain goals.
\item A domain consists of the actions that can be done and the restrictions on the actions.
\item A problem in the specified domain will consist of the initial conditions and the goal conditions for which a plan is a solution.
\end{itemize}
You will be dealing with the blocksworld domain which consists of playing with a set of blocks where you need to arrange the blocks into stacks.
You will have to come up with a plan for one blocksworld problem. 
 You get a base bonus of 50 cents.
 \begin{itemize}
\item If you come up with a successful plan your bonus compensation increases by \$1.
\item If your plan is unsuccessful, your bonus compensation decreases by 50 cents.
\item Random plan submissions will be rejected and the bonus compensation would not be provided for such submissions.
\end{itemize}
We recommend you to have a pen and paper to aid you in visualizing the domain whenever required. We will first look at how the blocksworld domain works and what actions can you do.

\mund{Study details for participants in the human baseline study}
In this study, you will be coming up with a plan that achieves certain goal conditions given some initial conditions.
\begin{itemize}
    \item A plan is a sequence of actions that achieve certain goals.
\item A domain consists of the actions that can be done and the restrictions on the actions.
\item A problem in the specified domain will consist of the initial conditions and the goal conditions for which a plan is a solution.
\end{itemize}
You will be dealing with the blocksworld domain which consists of playing with a set of blocks where you need to arrange the blocks into stacks.
You will have to come up with a plan for one blocksworld problem. 
 You get a base bonus of 50 cents.
 \begin{itemize}
\item If you come up with a successful plan your bonus compensation increases by 50 cents.
\item If your plan is unsuccessful, your bonus compensation decreases by 50 cents.
\item Random plan submissions will be rejected and the bonus compensation would not be provided for such submissions.
\end{itemize}
We recommend you to have a pen and paper to aid you in visualizing the domain whenever required. We will first look at how the blocksworld domain works and what actions can you do.
\subsubsection{Interface of the user study}
\begin{figure*}[ht]
    \centering
    \includegraphics[width=0.95\textwidth]{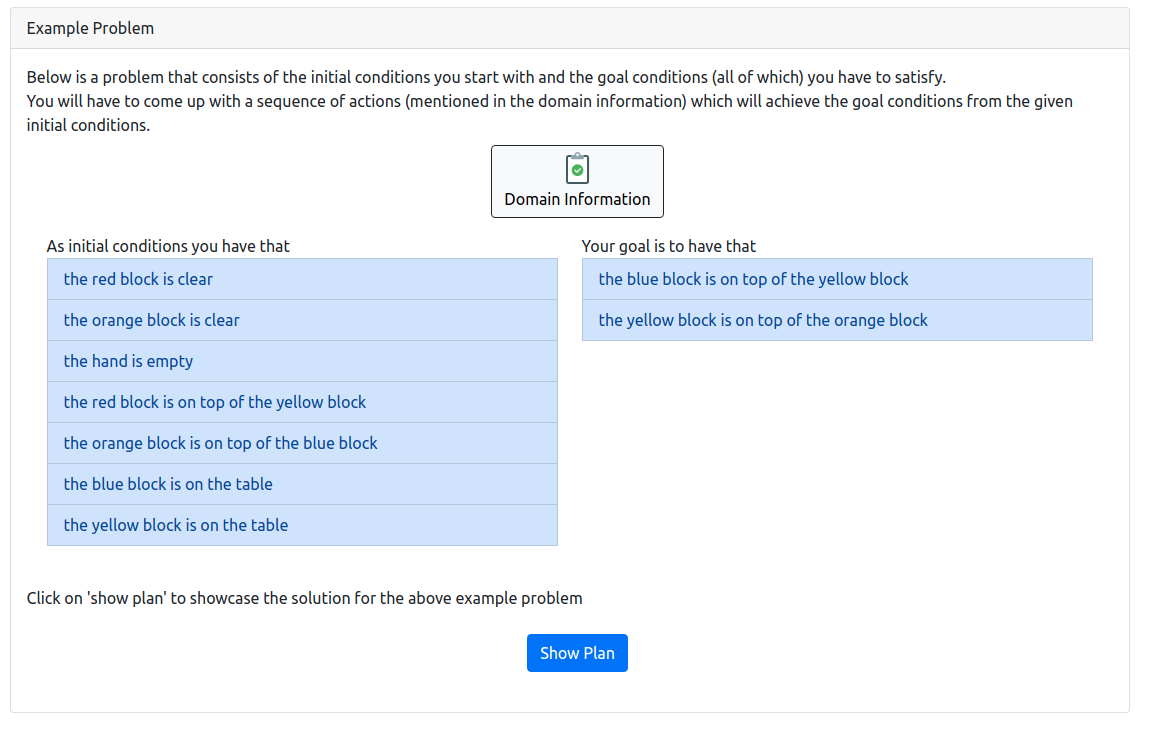}
    
    \caption{
The description of the example problem.
     }
    \label{fig:combined1}
\end{figure*}
\begin{figure*}[ht]
    \centering
    \includegraphics[width=0.95\textwidth]{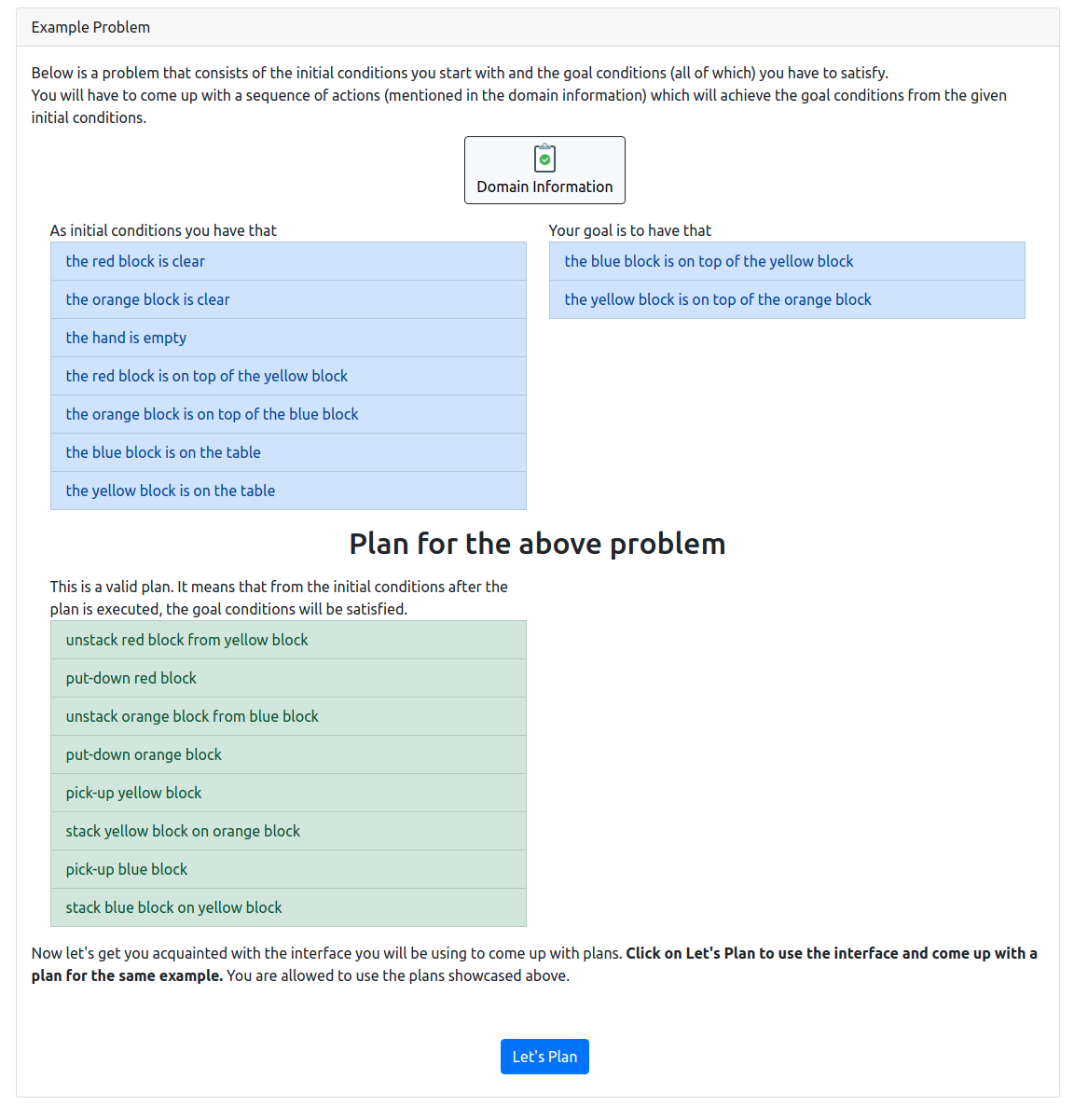}
    
    \caption{
The description of the example problem and showcasing the solution of the example problem.
     }
    \label{fig:combined2}
\end{figure*}
\begin{figure*}[ht]
    \centering
    \includegraphics[width=0.95\textwidth]{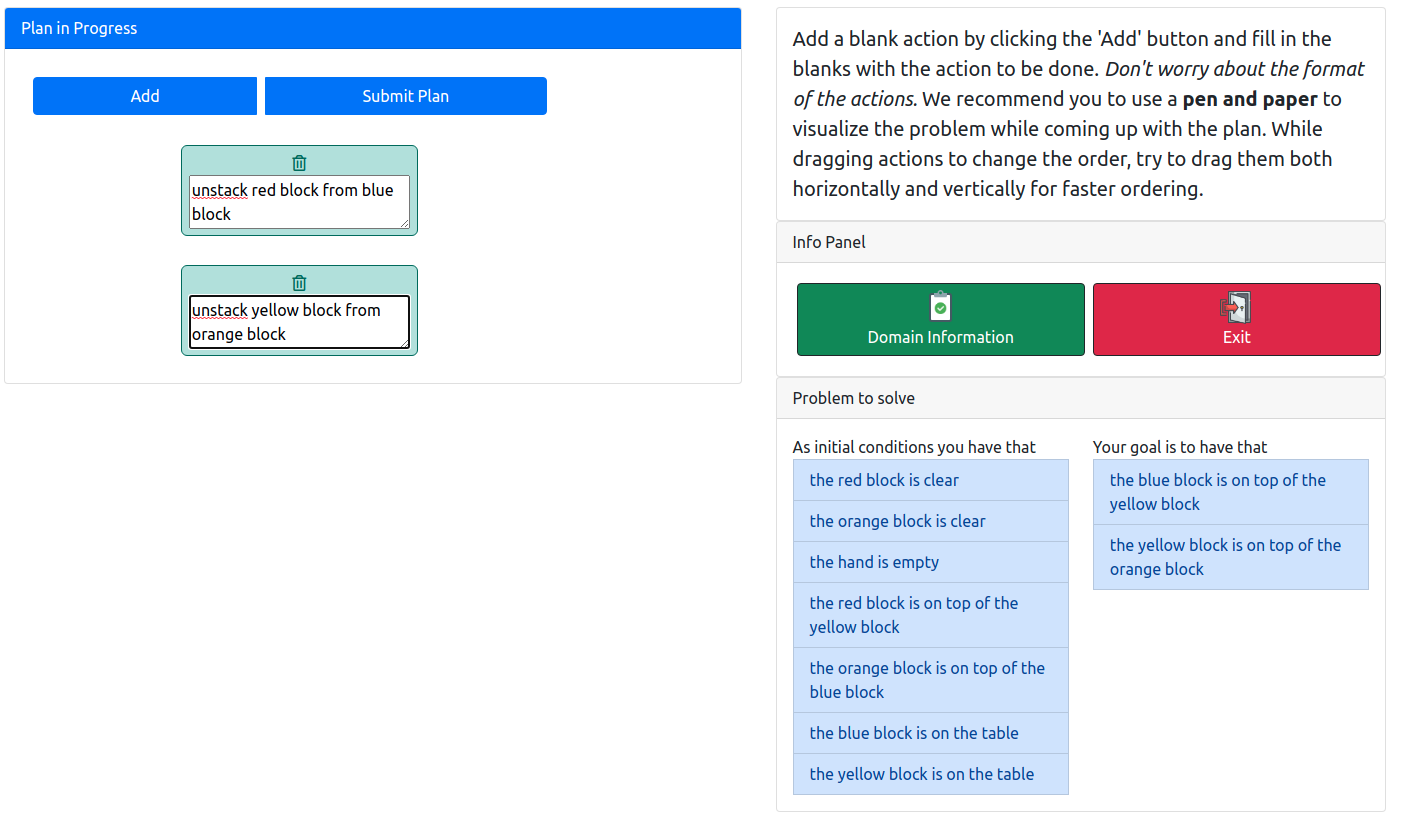}
    
    \caption{
Interface at the plan writing phase without LLM assistance.
     }
    \label{fig:combined3}
\end{figure*}
\begin{figure*}[ht]
    \centering
    \includegraphics[width=0.95\textwidth]{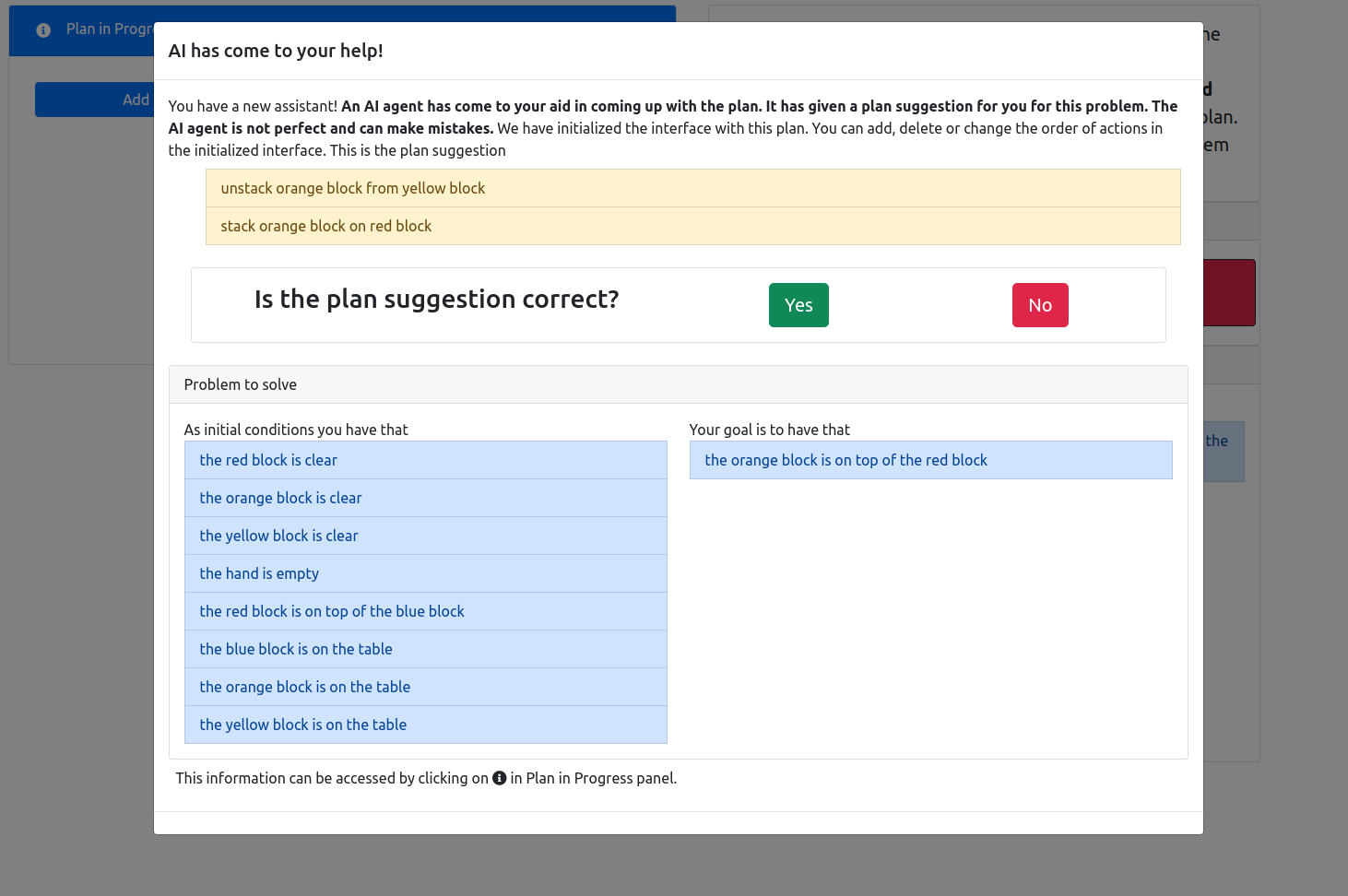}
    
    \caption{
Interface at plan writing phase with assistance from the LLM.
     }
    \label{fig:combined4}
\end{figure*}
\begin{figure*}[ht]
    \centering
    \includegraphics[width=0.95\textwidth]{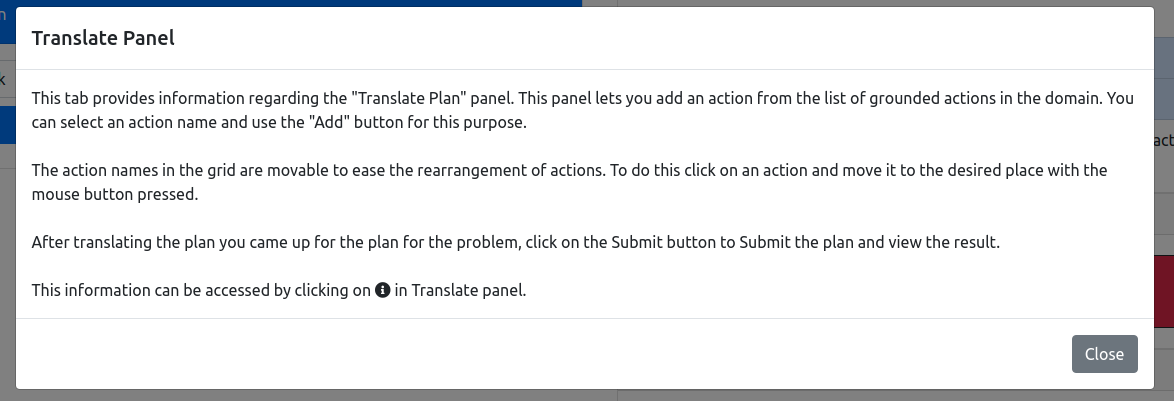}
    
    \caption{
Description of the translate panel.
     }
    \label{fig:combined5}
\end{figure*}
\begin{figure*}[ht]
    \centering
    \includegraphics[width=0.95\textwidth]{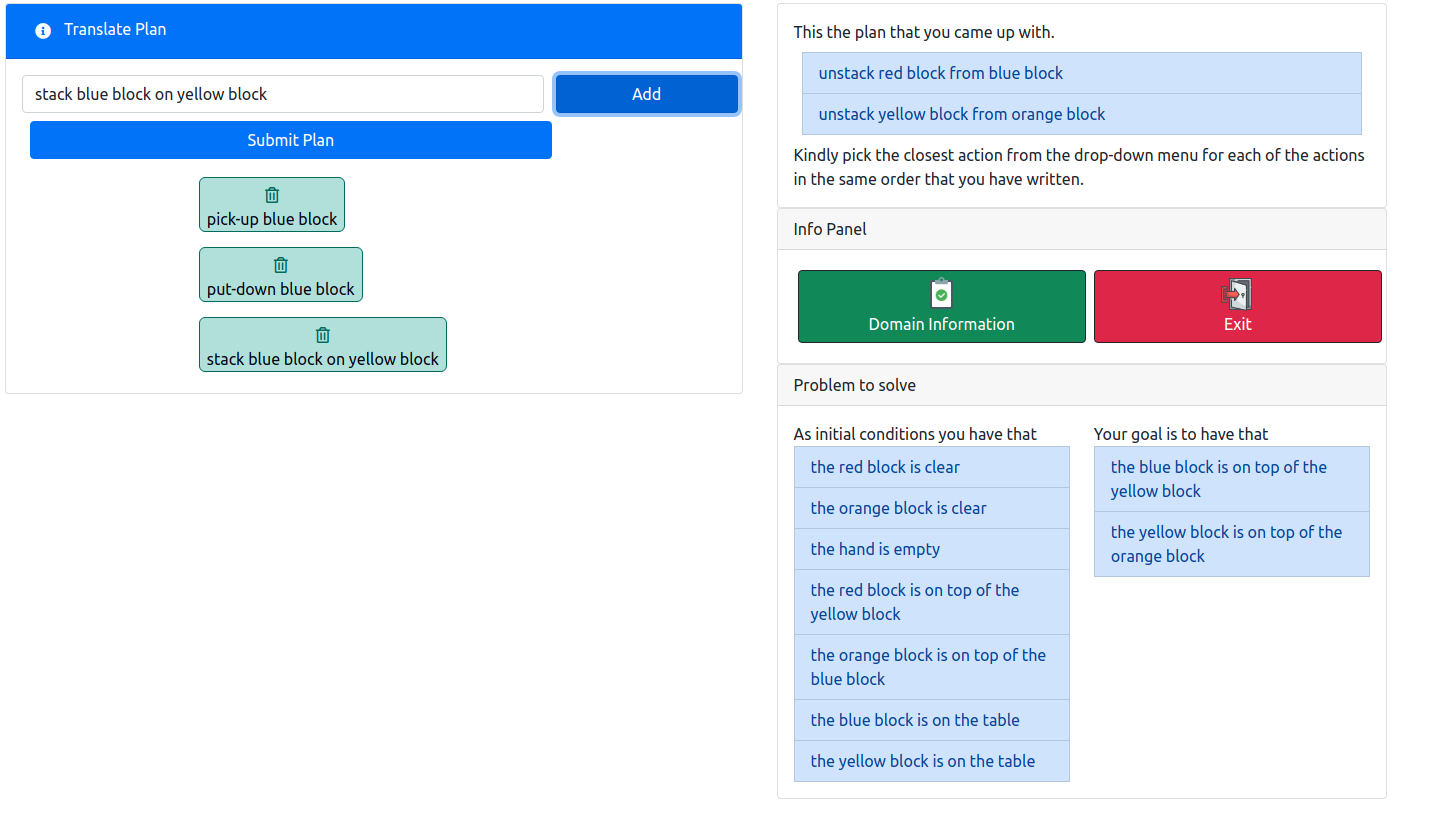}
    
    \caption{
Interface at the plan translation phase
     }
    \label{fig:combined6}
\end{figure*}

\begin{figure*}[ht]
    \centering
    \includegraphics[width=0.95\textwidth]{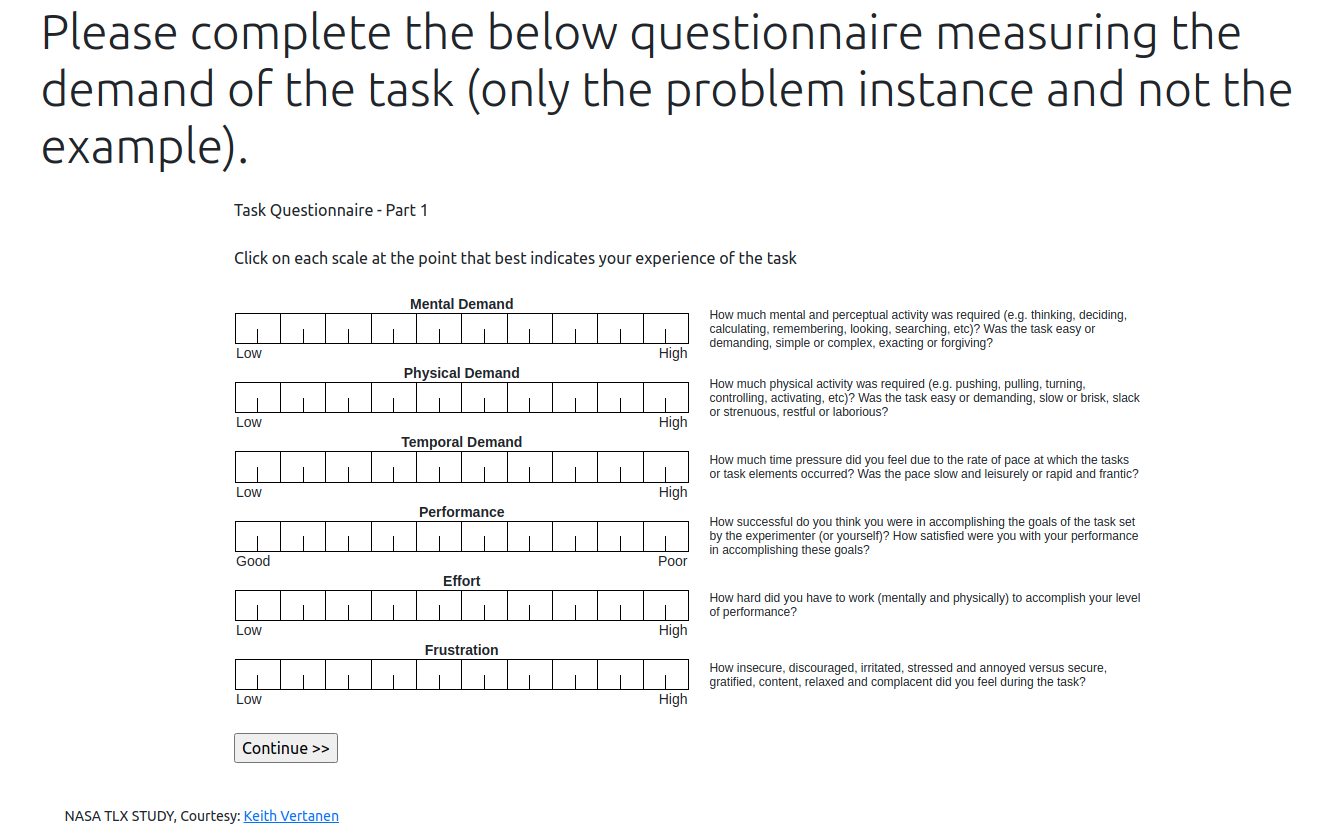}
    
    \caption{
NASA TLX assessment at the end of the study
     }
    \label{fig:combined7}
\end{figure*}

\end{document}

%% file: rao-intro.tex
\section{Introduction}
%Cite the ersatz natural science
It would be no exaggeration to say that transformer-based large language models (LLMs) have revolutionized the field of natural language processing (NLP). Kicked off by the advances presented by the GPT-x models developed by OpenAI \cite{radford2018improving}, these types of language models currently provide state-of-the-art performance in many of the standard NLP tasks. 
Although LLMs were originally developed mostly to do word sequence completion tasks, with no guarantees about the completion beyond its coherence, there have been increasing claims and anecdotal evidence that they have other {\em emergent capabilities} that are not normally associated with sequence completion.
%, including explaining jokes. 
Indeed, the hints of such emergent capabilities has started a veritable land rush, with researchers probing (prompting) and studying LLM behavior almost as if they were artificial organisms (c.f. \cite{kambhampati_2022}). Of particular interest to us in this paper is the thread of efforts that aim to investigate (and showcase) reasoning abilities of LLMs--including commonsense reasoning \cite{talmor2018commonsenseqa,sakaguchi2020winogrande, geva2021did}, logical reasoning \cite{bigbench}, and even ethical reasoning \cite{Jiang2021DelphiTM}. The macro-tenor of the drumbeat of these works has been suggesting that LLM’s are indeed capable of doing such kinds of reasoning \cite{kojima2022large, wei2022chain, chowdhery2022palm}.

%For example, there are many claims centered around the fact that GPT-3 may possess some form of reasoning ability \cite{li2021implicit}. Such sources generally assume that because the model learned from large amounts of real-world text, it may have acquired some approximation of simple reasoning. This sparked interest in evaluating the large language models on various reasoning tasks including common-sense reasoning 

One type of reasoning task that has been well studied in the AI community is planning and sequential decision making. At its simplest, planning involves developing a course of actions (policy) which when executed takes the agent to a desired state of the world. 
Planning has generally been studied primarily as an inference on world and reward models--whether specified by humans or learned by the agent by interacting with its world.
%If we train LLMs on transition data, they can certainly learn world models. 
%In the case of LLMs, 
In this paper, we are interested in seeing what planning abilities, if any, LLMs may already have, given their high capacity functions (with billions of tunable parameters) trained on web-scale corpora. Specifically, we are interested in answering two broad questions:
%We can further subdivide this investigation:

\begin{itemize}
\item How good are LLMs by themselves in generating and validating simple plans in commonsense planning tasks (of the type that humans are generally quite good at)? 
\item How good are LLMs in being a source of heuristic guidance for other agents--either AI planners or human planners--in their planning tasks?
\end{itemize}

Notice that in theory it is possible for LLMs to be very effective as idea generators for humans in the loop in computer-supported cooperative work scenarios, while themselves being very bad at generating plans that are guaranteed to be correct. This is especially likely because the chief power of LLMs comes from their pattern finding abilities than on first-principles simulations over world models. Compared to a planner that is guaranteed to be correct in a narrow set of domains, LLMs may likely be good at generating plausible (but not guaranteed to be correct) plan heuristics/suggestions in many more domains. 

To investigate these questions in a systematic rather than anecdotal manner, we start by developing a benchmark suite\footnote{Link to the github repo: \url{ https://github.com/karthikv792/gpt-plan-benchmark}}
%Thus, in this paper, we want to look at the ability of large language models to do reasoning about actions and change involving common-sense planning tasks. We develop a suite of benchmarks,
based on the kinds of domains employed in the International Planning Competition \cite{ipc}. The tasks in the benchmark suite are aimed to test a variety of plan generation and validation capabilities. 
To eliminate the subjective aspect of analysis that forms the core part of many earlier efforts on evaluating reasoning capabilities of LLMs,  we automate the evaluation by leveraging models and tools from the automated planning community. 
%and tools to generate the queries and validate the system's answers. 
While our primary interest is in plan generation, the test tasks themselves form a broad curriculum for evaluating LLM's capabilities of reasoning about actions and change.

The evaluation itself is done in three modes. In the first ``autonomous" mode, LLMs are used as stand alone, and we directly assess the quality and correctness of plans they generate. As we shall see, the results in the autonomous mode are pretty bleak. Only about 3\% of the plans that LLMs generate are actually executable without errors and reach their goals. We will show that the choice of the specific LLM (we experimented with two versions of GPT3 \cite{brown2020language, ouyang2022training} as well as BLOOM \cite{BLOOM}), as well as fine tuning seems to have little effect on this dismal performance. 
We also give a human baseline by presenting these tasks to human subjects (through IRB-approved studies) and evaluating the quality and correctness of their plans.  These results are \textit{substantially better} than those of LLMs--confirming that LLMs can't plan in autonomous mode. 

In the second ``heuristic" mode, the plans produced by LLMs are given as input to an automated planner working off of a correct domain model to check how easy it is to ``repair" the LLM plans to guarantee their correctness. 
%The results here are more promising--
Specifically we show that a well known automated planner called LPG \cite{gerevini2002lpg},  that uses local search to locate and remove flaws in a candidate plan to make it correct, is able to repair the LLM plans with relative ease. 
%Further an edit distance comparison of these corrected plans with the original LLM plans shows that the LLM plans are of the right form, albeit not being correct. 

In the third ``human-in-the-loop mode", the LLM plans are given to humans in the loop to see how it affects their ability to solve the bench mark tasks. The results here show modest improvements in the accuracy of the plans generated by humans when they start with LLM suggested plans. 

%This suggests that LLMs can be of use in  

Beyond our own initial studies, the goal of this work is to provide a systematic benchmark to evaluate the (evolving) planning capabilities of LLMs. To this end, we make the benchmark suite and the automated evaluation tools public to support further research.

%% file: llmplan.bbl
\begin{thebibliography}{10}

\bibitem{aeronautiques1998pddl}
Constructions Aeronautiques, Adele Howe, Craig Knoblock, ISI~Drew McDermott,
  Ashwin Ram, Manuela Veloso, Daniel Weld, David~Wilkins SRI, Anthony Barrett,
  Dave Christianson, et~al.
\newblock {PDDL| The Planning Domain Definition Language}.
\newblock {\em Technical Report, Tech. Rep.}, 1998.

\bibitem{ahn2022can}
Michael Ahn, Anthony Brohan, Noah Brown, Yevgen Chebotar, Omar Cortes, Byron
  David, Chelsea Finn, Keerthana Gopalakrishnan, Karol Hausman, Alex Herzog,
  et~al.
\newblock Do as i can, not as i say: Grounding language in robotic affordances.
\newblock {\em arXiv preprint arXiv:2204.01691}, 2022.

\bibitem{brown2020language}
Tom~B Brown, Benjamin Mann, Nick Ryder, Melanie Subbiah, Jared Kaplan, Prafulla
  Dhariwal, Arvind Neelakantan, Pranav Shyam, Girish Sastry, Amanda Askell,
  et~al.
\newblock Language models are few-shot learners.
\newblock {\em Advances in neural information processing systems},
  33:1877--1901, 2020.

\bibitem{chen2021decision}
Lili Chen, Kevin Lu, Aravind Rajeswaran, Kimin Lee, Aditya Grover, Misha
  Laskin, Pieter Abbeel, Aravind Srinivas, and Igor Mordatch.
\newblock Decision transformer: Reinforcement learning via sequence modeling.
\newblock {\em Advances in neural information processing systems},
  34:15084--15097, 2021.

\bibitem{chowdhery2022palm}
Aakanksha Chowdhery, Sharan Narang, Jacob Devlin, Maarten Bosma, Gaurav Mishra,
  Adam Roberts, Paul Barham, Hyung~Won Chung, Charles Sutton, Sebastian
  Gehrmann, et~al.
\newblock Palm: Scaling language modeling with pathways.
\newblock {\em arXiv preprint arXiv:2204.02311}, 2022.

\bibitem{cobbe2021training}
Karl Cobbe, Vineet Kosaraju, Mohammad Bavarian, Jacob Hilton, Reiichiro Nakano,
  Christopher Hesse, and John Schulman.
\newblock Training verifiers to solve math word problems.
\newblock {\em arXiv preprint arXiv:2110.14168}, 2021.

\bibitem{cummings2017automation}
Mary~L Cummings.
\newblock Automation bias in intelligent time critical decision support
  systems.
\newblock In {\em Decision making in aviation}, pages 289--294. Routledge,
  2017.

\bibitem{du2021glam}
Nan Du, Yanping Huang, Andrew~M Dai, Simon Tong, Dmitry Lepikhin, Yuanzhong Xu,
  Maxim Krikun, Yanqi Zhou, Adams~Wei Yu, Orhan Firat, et~al.
\newblock {GLaM: Efficient Scaling of Language Models with Mixture-of-Experts}.
\newblock pages 5547--5569, 2022.

\bibitem{gerevini2002lpg}
Alfonso Gerevini and Ivan Serina.
\newblock Lpg: A planner based on local search for planning graphs with action
  costs.
\newblock In {\em AIPS}, volume~2, pages 281--290, 2002.

\bibitem{geva2021did}
Mor Geva, Daniel Khashabi, Elad Segal, Tushar Khot, Dan Roth, and Jonathan
  Berant.
\newblock Did aristotle use a laptop? a question answering benchmark with
  implicit reasoning strategies.
\newblock {\em Transactions of the Association for Computational Linguistics},
  9:346--361, 2021.

\bibitem{helmert2006fast}
Malte Helmert.
\newblock The fast downward planning system.
\newblock {\em Journal of Artificial Intelligence Research}, 26:191--246, 2006.

\bibitem{hoffmann2022training}
Jordan Hoffmann, Sebastian Borgeaud, Arthur Mensch, Elena Buchatskaya, Trevor
  Cai, Eliza Rutherford, Diego de~Las Casas, Lisa~Anne Hendricks, Johannes
  Welbl, Aidan Clark, et~al.
\newblock Training compute-optimal large language models.
\newblock {\em arXiv preprint arXiv:2203.15556}, 2022.

\bibitem{howey2004val}
Richard Howey, Derek Long, and Maria Fox.
\newblock {VAL: Automatic plan validation, continuous effects and mixed
  initiative planning using PDDL}.
\newblock In {\em 16th IEEE International Conference on Tools with Artificial
  Intelligence}, pages 294--301. IEEE, 2004.

\bibitem{huang2022language}
Wenlong Huang, Pieter Abbeel, Deepak Pathak, and Igor Mordatch.
\newblock Language models as zero-shot planners: Extracting actionable
  knowledge for embodied agents.
\newblock {\em arXiv preprint arXiv:2201.07207}, 2022.

\bibitem{ipc}
IPC.
\newblock International planning competition, 1998.

\bibitem{Jiang2021DelphiTM}
Liwei Jiang, Jena~D. Hwang, Chandrasekhar Bhagavatula, Ronan~Le Bras, Maxwell
  Forbes, Jon Borchardt, Jenny Liang, Oren Etzioni, Maarten Sap, and Yejin
  Choi.
\newblock {Delphi: Towards Machine Ethics and Norms}.
\newblock {\em ArXiv}, abs/2110.07574, 2021.

\bibitem{kambhampati_2022}
Subbarao Kambhampati.
\newblock {AI as (an Ersatz) Natural Science?}
\newblock
  https://cacm.acm.org/blogs/blog-cacm/261732-ai-as-an-ersatz-natural-science/fulltext,
  Jun 2022.

\bibitem{kojima2022large}
Takeshi Kojima, Shixiang~Shane Gu, Machel Reid, Yutaka Matsuo, and Yusuke
  Iwasawa.
\newblock {Large Language Models are Zero-Shot Reasoners}.
\newblock {\em arXiv preprint arXiv:2205.11916}, 2022.

\bibitem{ling2017program}
Wang Ling, Dani Yogatama, Chris Dyer, and Phil Blunsom.
\newblock {Program induction by rationale generation: Learning to solve and
  explain algebraic word problems}.
\newblock {\em arXiv preprint arXiv:1705.04146}, 2017.

\bibitem{nguyen2012generating}
Tuan~Anh Nguyen, Minh Do, Alfonso~Emilio Gerevini, Ivan Serina, Biplav
  Srivastava, and Subbarao Kambhampati.
\newblock Generating diverse plans to handle unknown and partially known user
  preferences.
\newblock {\em Artificial Intelligence}, 190:1--31, 2012.

\bibitem{ouyang2022training}
Long Ouyang, Jeff Wu, Xu~Jiang, Diogo Almeida, Carroll~L Wainwright, Pamela
  Mishkin, Chong Zhang, Sandhini Agarwal, Katarina Slama, Alex Ray, et~al.
\newblock Training language models to follow instructions with human feedback.
\newblock {\em arXiv preprint arXiv:2203.02155}, 2022.

\bibitem{patel2021nlp}
Arkil Patel, Satwik Bhattamishra, and Navin Goyal.
\newblock {Are NLP Models really able to Solve Simple Math Word Problems?}
\newblock {\em arXiv preprint arXiv:2103.07191}, 2021.

\bibitem{puig2018virtualhome}
Xavier Puig, Kevin Ra, Marko Boben, Jiaman Li, Tingwu Wang, Sanja Fidler, and
  Antonio Torralba.
\newblock {Virtualhome: Simulating household activities via programs}.
\newblock In {\em Proceedings of the IEEE Conference on Computer Vision and
  Pattern Recognition}, pages 8494--8502, 2018.

\bibitem{radford2018improving}
Alec Radford, Karthik Narasimhan, Tim Salimans, and Ilya Sutskever.
\newblock Improving language understanding by generative pre-training.
\newblock 2018.

\bibitem{rae2021scaling}
Jack~W Rae, Sebastian Borgeaud, Trevor Cai, Katie Millican, Jordan Hoffmann,
  Francis Song, John Aslanides, Sarah Henderson, Roman Ring, Susannah Young,
  et~al.
\newblock Scaling language models: Methods, analysis \& insights from training
  gopher.
\newblock {\em arXiv preprint arXiv:2112.11446}, 2021.

\bibitem{reed2022generalist}
Scott Reed, Konrad Zolna, Emilio Parisotto, Sergio~Gomez Colmenarejo, Alexander
  Novikov, Gabriel Barth-Maron, Mai Gimenez, Yury Sulsky, Jackie Kay,
  Jost~Tobias Springenberg, et~al.
\newblock A generalist agent.
\newblock {\em arXiv preprint arXiv:2205.06175}, 2022.

\bibitem{sakaguchi2020winogrande}
Keisuke Sakaguchi, Ronan Le~Bras, Chandra Bhagavatula, and Yejin Choi.
\newblock Winogrande: An adversarial winograd schema challenge at scale.
\newblock In {\em Proceedings of the AAAI Conference on Artificial
  Intelligence}, volume~34, pages 8732--8740, 2020.

\bibitem{BLOOM}
Big Science.
\newblock {BigScience Large Open-science Open-access Multilingual Language
  Model}.
\newblock \url{https://huggingface.co/bigscience/bloom}, 2022.

\bibitem{silver2022pddl}
Tom Silver, Varun Hariprasad, Reece~S Shuttleworth, Nishanth Kumar, Tom{\'a}s
  Lozano-P{\'e}rez, and Leslie~Pack Kaelbling.
\newblock {PDDL} planning with pretrained large language models.
\newblock In {\em NeurIPS 2022 Foundation Models for Decision Making Workshop},
  2022.

\bibitem{smith2022using}
Shaden Smith, Mostofa Patwary, Brandon Norick, Patrick LeGresley, Samyam
  Rajbhandari, Jared Casper, Zhun Liu, Shrimai Prabhumoye, George Zerveas,
  Vijay Korthikanti, et~al.
\newblock Using deepspeed and megatron to train megatron-turing nlg 530b, a
  large-scale generative language model.
\newblock {\em arXiv preprint arXiv:2201.11990}, 2022.

\bibitem{bigbench}
Aarohi Srivastava, Abhinav Rastogi, Abhishek Rao, Abu Awal~Md Shoeb, Abubakar
  Abid, Adam Fisch, Adam~R Brown, Adam Santoro, Aditya Gupta, Adri{\`a}
  Garriga-Alonso, et~al.
\newblock Beyond the imitation game: Quantifying and extrapolating the
  capabilities of language models.
\newblock {\em arXiv preprint arXiv:2206.04615}, 2022.

\bibitem{srivastava2011qualitative}
Siddharth Srivastava, Shlomo Zilberstein, Neil Immerman, and Hector Geffner.
\newblock Qualitative numeric planning.
\newblock In {\em Proceedings of the AAAI Conference on Artificial
  Intelligence}, volume~25, pages 1010--1016, 2011.

\bibitem{talmor2018commonsenseqa}
Alon Talmor, Jonathan Herzig, Nicholas Lourie, and Jonathan Berant.
\newblock Commonsenseqa: A question answering challenge targeting commonsense
  knowledge.
\newblock {\em arXiv preprint arXiv:1811.00937}, 2018.

\bibitem{thoppilan2022lamda}
Romal Thoppilan, Daniel De~Freitas, Jamie Hall, Noam Shazeer, Apoorv
  Kulshreshtha, Heng-Tze Cheng, Alicia Jin, Taylor Bos, Leslie Baker, Yu~Du,
  et~al.
\newblock {LaMDA: Language Models for Dialog Applications}.
\newblock {\em arXiv preprint arXiv:2201.08239}, 2022.

\bibitem{wei2022chain}
Jason Wei, Xuezhi Wang, Dale Schuurmans, Maarten Bosma, Ed~Chi, Quoc Le, and
  Denny Zhou.
\newblock Chain of thought prompting elicits reasoning in large language
  models.
\newblock {\em arXiv preprint arXiv:2201.11903}, 2022.

\bibitem{zhang2022paradox}
Honghua Zhang, Liunian~Harold Li, Tao Meng, Kai-Wei Chang, and Guy Van~den
  Broeck.
\newblock On the paradox of learning to reason from data.
\newblock {\em arXiv preprint arXiv:2205.11502}, 2022.

\bibitem{zhang2022opt}
Susan Zhang, Stephen Roller, Naman Goyal, Mikel Artetxe, Moya Chen, Shuohui
  Chen, Christopher Dewan, Mona Diab, Xian Li, Xi~Victoria Lin, et~al.
\newblock {Opt: Open pre-trained transformer language models}.
\newblock {\em arXiv preprint arXiv:2205.01068}, 2022.

\bibitem{zhuo2020discovering}
Hankz~Hankui Zhuo, Yantian Zha, Subbarao Kambhampati, and Xin Tian.
\newblock Discovering underlying plans based on shallow models.
\newblock {\em ACM Transactions on Intelligent Systems and Technology (TIST)},
  11(2):1--30, 2020.

\end{thebibliography}
